\documentclass{article}
\usepackage{graphicx}

\PassOptionsToPackage{numbers}{natbib} % <-- Add this line
\usepackage[preprint]{acl} 
\usepackage[utf8]{inputenc} % allow utf-8 input
\usepackage[T1]{fontenc}    % use 8-bit T1 fonts
\usepackage{hyperref}       % hyperlinks
\usepackage{url}            % simple URL typesetting
\usepackage{booktabs}       % professional-quality tables
\usepackage{amsfonts}       % blackboard math symbols
\usepackage{nicefrac}       % compact symbols for 1/2, etc.
\usepackage{microtype}      % microtypography
\usepackage[table,xcdraw]{xcolor}
\usepackage{booktabs} % for tables
\usepackage{multirow} % for tables
\usepackage{makecell} % for line breaks in cells in tables
\usepackage{listings} % For formatting prompts in the appendix
\usepackage[cache=false]{minted}
\usepackage[most]{tcolorbox}
\tcbuselibrary{minted}
\usepackage{pifont}
\usepackage{tabularx}

\newcommand{\cmark}{\textcolor{green!60!black}{\ding{51}}} % check
\newcommand{\xmark}{\textcolor{red}{\ding{55}}}

\title{Continual Learning of Domain Knowledge from Human Feedback in Text-to-SQL}

\author{
  Thomas Cook$^{\star}$ \quad
  Kelly Patel \quad
  Sivapriya Vellaichamy \\
  \textbf{Udari Madhushani Sehwag}$^{\mathsection}$ \quad
  \textbf{Saba Rahimi} \quad
  \textbf{Zhen Zeng} \quad
  \textbf{Sumitra Ganesh} \\
  JPMorganChase \\
}

\begin{document}

\maketitle
\begingroup
\renewcommand\thefootnote{}\footnotetext{
$^{\star}$ Corresponding author \par \texttt{thomas.cook@jpmchase.com} \par
$^{\mathsection}$ All research work was conducted while the author was at JPMorganChase.
}
\endgroup

\begin{abstract}
Large Language Models (LLMs) can generate SQL queries from natural language questions but struggle with database-specific schemas and tacit domain knowledge. We introduce a framework for continual learning from human feedback in text-to-SQL, where a learning agent receives natural language feedback to refine queries and distills the revealed knowledge for reuse on future tasks. This distilled knowledge is stored in a structured memory, enabling the agent to improve execution accuracy over time. We design and evaluate multiple variations of a learning agent architecture that vary in how they capture and retrieve past experiences. Experiments on the BIRD benchmark Dev set show that memory-augmented agents, particularly the Procedural Agent, achieve significant accuracy gains and error reduction by leveraging human-in-the-loop feedback. Our results highlight the importance of transforming tacit human expertise into reusable knowledge, paving the way for more adaptive, domain-aware text-to-SQL systems that continually learn from a human-in-the-loop.
\end{abstract}

\section{Introduction}
While AI Agents, with Large Language Models (LLMs) as backbones, have shown advanced capabilities performing complex tasks, they still require humans to prepare domain specific information to generate accurate outputs. In many applications, this information may not be readily available in formal documentation or procedures, but rather exists as tacit knowledge, disseminated through informal channels such as personal experience and word-of-mouth communication, where experienced individuals convey insights, unwritten rules, and practical know-how. This unstructured, experiential knowledge is often difficult to anticipate and formally integrate, posing a significant challenge for deploying AI systems in practice.

This challenge is especially critical in the text-to-SQL domain. Although LLMs are highly adept at generating syntactically correct SQL queries, they often lack a nuanced understanding of database-specific schema and semantics essential for generating executable queries. Complex databases encompass a vast and elaborate set of tables, relationships, and constraints, and demand a level of domain expertise that is seldom captured during pretraining or simple prompt tuning. Consequently, there exists a marked gap between the general query-generation capabilities of these models and the specialized requirements of practical usage.

By focusing on continual learning through human feedback, we aim to enable a learning agent to effectively operate in a cold start regime, adapting and improving its performance through ongoing interaction to complete a variety of tasks. We endow the learning agent to iteratively refine its output based on provided natural language feedback. At the conclusion of each interaction, the learning agent reflects on the interaction to extract the acquired domain knowledge. By storing this extracted information along with previous input-output pairs, the learning agent has a growing store of examples and prior experiences to leverage for in-context learning for future tasks, enabling continual learning. 

Given these challenges, we propose a novel framework that leverages continual learning through human-in-the-loop feedback to iteratively refine LLM-generated SQL queries. Our approach centers on directly engaging a human expert to bridge the domain knowledge gap, extracting and distilling tacit knowledge, breaking it down to the procedural level and making use of subtask logic during the interaction between a learning agent and the human. To facilitate continual learning, we provide the learning agent with a dynamic memory, which empirically improves performance as more interactions with the human occur. The key contributions of our work are as follows:
\begin{itemize}
    \item \textbf{Direct Integration of Tacit Knowledge}: We demonstrate how expert interactions can be used to capture subtle, domain-specific insights not captured through input-output example pairs, thereby enhancing the operational accuracy of text-to-SQL systems.

    \item \textbf{Continual Learning via Memory Augmentation}: By incorporating a structured memory mechanism that stores observed natural language queries, corresponding SQL outputs, and expert feedback distilled into procedural knowledge, our framework enables the agent to improve its performance in a cold start setting. We show that breaking down the feedback into procedural steps enables the agent to generalize concepts across similar tasks as well as provide more granular improvements to SQL generation which significantly boosts performance.

    \item \textbf{Comparative Agent Architectures}: We systematically construct and evaluate agent designs by incrementally enabling distinct memory types and reasoning. Starting from a baseline in-context learning setting, we incrementally develop and ablate components leading to our proposed \emph{Procedural Agent}, analyzing how each architectural enhancement affects SQL query generation accuracy and error reduction.
\end{itemize}

By addressing the intrinsic limitations of LLMs in capturing domain-specific nuances, our work provides a pathway for developing AI agents that continually learn. In the sections that follow, we formalize the problem setting, detail the architecture of our proposed agents, and present empirical evaluations demonstrating the performance gains achieved through our continual learning approach.

\section{Technical Preliminaries}

\begin{figure*}[h]
  \centering
  \includegraphics[width=\textwidth]{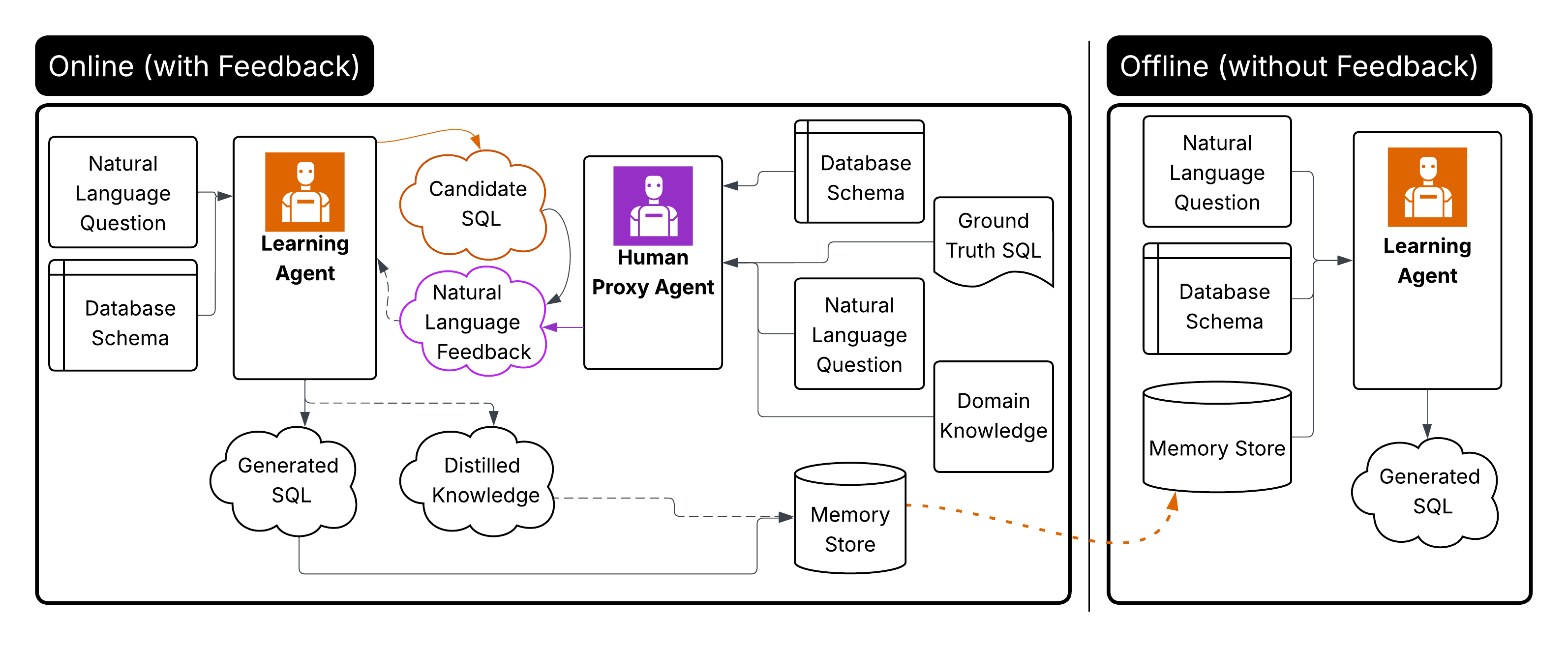}
  \caption{Overview of the two settings used for testing. In the online setting (left), the Learning Agent takes a natural language question and schema as input, generates a candidate SQL query, and iteratively refines it based on feedback from a human proxy agent. The final correct SQL and the distilled knowledge from this interaction are then stored in memory. In the offline phase (right), the Learning Agent generates SQL queries based on the natural language question, database schema, and the memory store (if available), without further feedback. The Learning Agent progressively improves execution accuracy by augmenting its memory.}
  \label{fig:test_setup}
\end{figure*}

\subsection{Problem Setting}
We consider the setting where we observe a streaming sequence of natural language questions (NLQ), $\{X_1, X_2, \dots\}$, with corresponding ground-truth SQL queries, $\{Y_1, Y_2, \dots\}$. We define a Learning Agent, $\mathcal{L}$, that takes as inputs the natural language question $X_t$ and a historical memory $M_t$, and outputs a generated SQL query, $\hat{Y}_t$; $\mathcal{L} : \{ \mathcal{X} \} \times \{ \mathcal{M} \}\mapsto \{\mathcal{Y}\}$. Feedback, $F_t$, is also provided on intermediate generated SQL queries, meaning that the historical memory at time $T$ depends only on $\{X_t\}_{t=1}^{T-1}$, $\{Y_t\}_{t=1}^{T-1}$, and $\{F_t\}_{t=1}^{T-1}$. 

We evaluate the agent using \emph{execution accuracy}, which asses whether the execution result of the generated SQL query is identical to that of the ground-truth\footnote{The generated query can differ syntactically, but produce the same output.}. We denote the accuracy with a binary variable $Z_t$. 

To assess an agent’s ability to continually learn, we measure how its performance on a holdout set, $\{{X'_i}, Y'_i\}_{i=1}^k$, improves as it accumulates more experience. Over time, as the agent processes more queries and builds its accumulated experience, we monitor the proportion of queries for which the execution output matches that of the ground truth, $\bar{Z_T} = k^{-1}\sum_{i=1}^{k} Z'_i$. Improvement in this performance metric indicates that the agent is learning effectively.

\subsection{Related Work}

\subsubsection{Context Augmentation for Continual Learning}
\citet{brown2020languagemodelsfewshotlearners} established that large-scale LLMs can adapt to new tasks simply by conditioning on a few input‑output examples in their prompt (in‑context learning), without parameter updates. \citet{madaan-etal-2022-memory} demonstrate how this phenomena naturally lends itself to continual learning in our problem setting, by continuously enriching the prompt context over time with new, more relevant experiences effectively allow an agent to \emph{learn} within the context provided. 
Recent agentic systems have formalized this idea. \citet{zheng2024synapse} propose Synapse, where agents store entire state‑action trajectories as exemplars and retrieve them in context to improve performance on future tasks in multi-step control settings, thus continually learning from past experiences. 
Similarly, Agent Workflow Memory \citep{wang2025agent} extracts reusable workflows from past executions and includes these workflows in the agent's context to guide future task solving.  \citet{liu2025contextual} introduce Contextual Experience Replay (CER), wherein agents accumulate past experiences in a memory buffer and replay them through the prompt to better adapt to new tasks without any model fine-tuning. 
Collectively, these works illustrate how in‑context memory augmentation enables agents to continually improve their decision-making by grounding future prompts in their accumulated experiences, however, these methods do not explicitly considering the extraction of tacit knowledge from human feedback, which we believe is critical in the text-to-SQL domain and is the primary focus of our work.

\subsubsection{Text-to-SQL}
LLMs achieve state-of-the-art results on the BIRD \cite{li2024can} and SPIDER \cite{yu-etal-2018-spider} leaderboards, often without fine-tuning \cite{shkapenyuk2025automaticmetadataextractiontexttosql,pourreza2025chasesql}. Current pipelines rely on query decomposition \citep{wang-etal-2025-mac}, schema linking \citep{shkapenyuk2025automaticmetadataextractiontexttosql}, in-context examples \citep{pourreza2025chasesql}, multiple candidate generations \citep{pourreza2025chasesql, liu2025xiyansqlnovelmultigeneratorframework, lee2024mcssqlleveragingmultipleprompts}, and chain-of-thought reasoning \citep{pourreza2025chasesql}. Incorporating feedback and domain adaptation have been explored via synthetic examples \cite{tian2025SQLSynth}, reflective notebooks \cite{chu2024leveragingpriorexperienceexpandable}, and fine-tuning \cite{chen2023yesterday}. Our method extends these works by distilling knowledge from trajectories that capture errors, natural language feedback from a human, and the revisions made. Our learning agent organizes this knowledge across multiple memory stores at different granularities, supporting continual learning specifically for chain-of-thought style procedural reasoning.

\subsection{Human Feedback in Text-to-SQL} Adapting to new databases or evolving schemas remains a significant challenge in the text-to-SQL domain. \citet{tian2025SQLSynth} proposed \textsc{SQLSynth}, an LLM-based, human-in-the-loop annotation method designed to generate a synthetic dataset for fine-tuning. This approach allows humans to manually update the correctness and alignment of generated SQL queries. However, their solution is presented as a static operation rather than a continual learning process. In contrast, \citet{tian-etal-2023-interactive} consider human feedback in the form of editing step-by-step explanations, yet they do not explore the reuse of this feedback for future steps. Addressing this gap, \citet{Menon2025FISQL} introduced \textsc{FISQL}, which refines SQL queries by classifying natural language feedback from users and making minimal prompt adjustments, thereby enhancing the adaptability and accuracy of the text-to-SQL process, which is what we build off in our work.

\subsubsection{Continual Learning in Text-to-SQL} The aforementioned works do not explicitly consider the task of continual learning. Our work most closely resembles the work of \citet{chu2024leveragingpriorexperienceexpandable}, who propose \textsc{LPE-SQL}, which builds auxiliary notebooks that contain previous NLQ-SQL pairs augmented with reasoning paths and self-reflection generated tips. Their framework does not consider the use of natural language feedback from an expert, but instead uses a self reflection mechanism, that is based on compiler hints and ground-truth SQL query. A key difference, is that in our set up, the refinement process is carried out to completion (or until a maximum number of steps is reached), producing a trajectory that details key mistakes and the necessary revisions. This trajectory-based distillation preserves actionable insights that can be reused for subsequent queries, rather than abstract reflections that may be vague. Moreso, our Procedural Agent makes use of multiple memory stores, which contain distilled knowledge at different granularities. \citet{chen2023yesterday} consider continual learning in Text-to-SQL and propose a combination of semi-supervised learning and and memory-replay continual learning for an LSTM model. This methodology requires access to model weights and may be cumbersome for use on large foundation models.

\section{Agent Design}

\subsection{Learning Agent}
We focus on a learning agent which relies on an LLM as its backbone and incorporates previous experiences through prompt augmentation. Specifically, we use GPT-4o as the LLM backbone \citep{openai2024gpt4o}, and use text-embedding-ada-002 as the embedding model used for retrieval tasks \cite{openai2022ada002}.

We implement a baseline modeled after the GPT-4 setup reported on the BIRD leaderboard \cite{li2024can}, which uses a templated prompt containing basic instructions, database schema, and in-context examples. While recent text-to-SQL systems incorporate increasingly complex architectures, our goal is to isolate the role of continual learning. We therefore focus on the agent’s memory and reasoning components rather than incorporating additional advances that could confound the analysis. With this agent, we can ablate the memory architecture by varying the available granularities, alongside enabling procedural reasoning for SQL construction. We denote a non-procedural agent with NP and a procedural agent with P. Detailed prompts and tool definitions are provided in Appendix~\ref{appdx:prompts}.

\subsubsection{Memory Design}

\begin{figure}[h!]
    \centering
    \includegraphics[width=\linewidth]{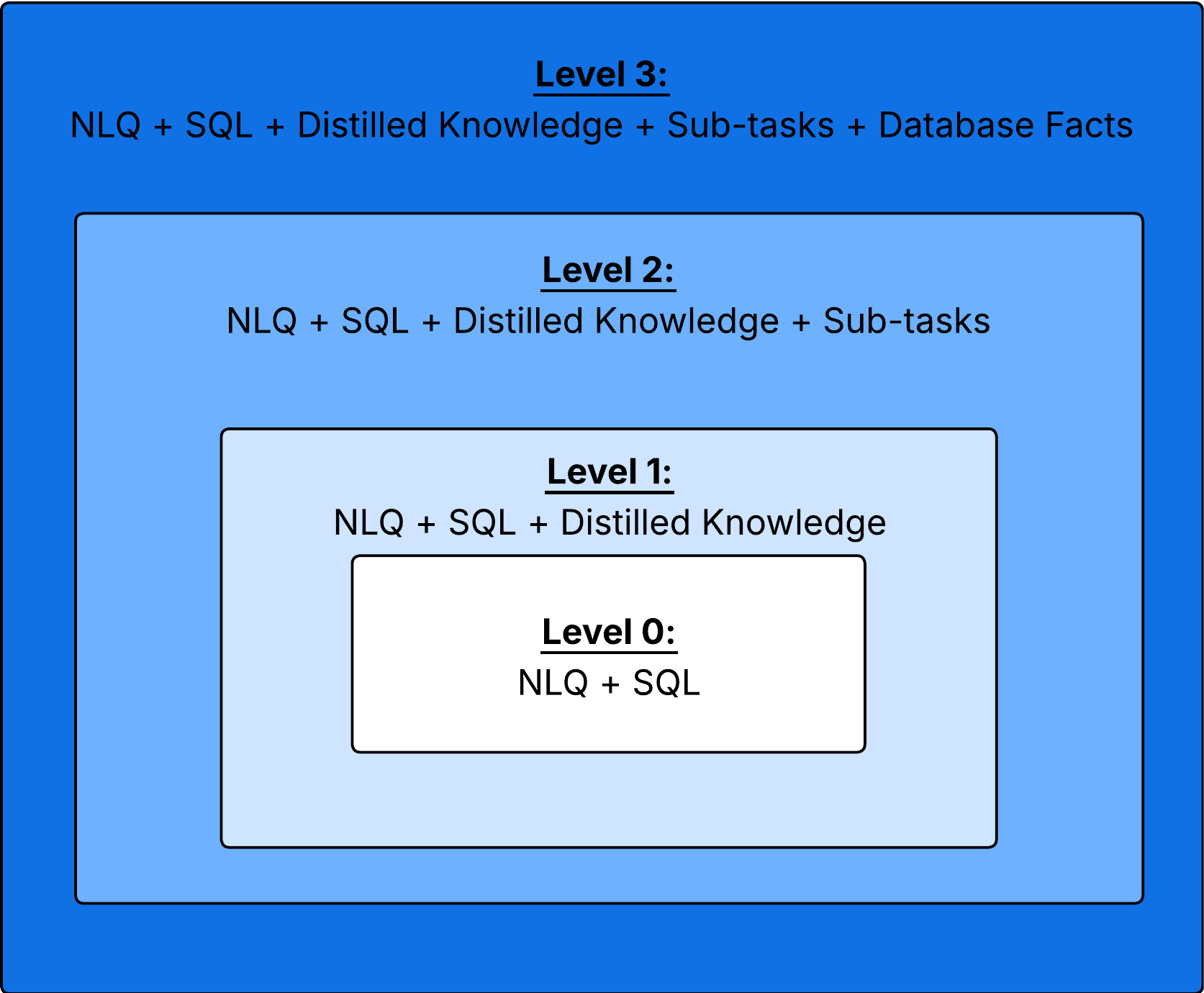}
    \caption{Levels of Memory Granularity}
    \label{fig:mem_hierarchy}
\end{figure}

The Learning Agent retrieves relevant knowledge from a memory store, organized into four levels of granularity, shown in Figure~\ref{fig:mem_hierarchy}. Level 0 contains only NLQ–SQL Query pairs from previous successful runs and serves as a baseline representation for in-context learning. Level 1 extends this with knowledge distilled from agent-human interactions, emphasizing tacit knowledge revealed in the human feedback. Details on the distillation process are provided in Section~\ref{sec:distill} and Appendix~\ref{appdx:prompts}. Level 2 further decomposes tasks into subtasks, storing each with affiliated SQL snippets in a separate memory store that the agent can access flexibly during SQL generation. Finally, Level 3 incorporates a memory store for database facts, capturing knowledge about the database that is not directly represented as SQL snippets. Examples of the contents of each memory store are provided in Appendix~\ref{appdx:trajectories}.

\subsubsection{Knowledge Distillation}\label{sec:distill}

Upon successfully generating a SQL query, the Learning Agent distills knowledge from the interaction trajectory through a guided chain-of-thought reasoning process. Trajectories may be long, with multiple errors and iterative revisions, so the agent first explicitly identifies its mistakes, organizing the key failures to the end of the LLM's context to prevent omitting any errors. Next, the agent isolates the information provided by the human that corrected these mistakes. This step captures tacit knowledge rather than generic SQL skills which the agent already possesses. Finally, the agent rephrases this knowledge to a broader context, generalizing the knowledge beyond the scope of the current question, to enable reuse on other questions. The resulting distilled knowledge is stored in Level 1 memory, while the learning agent uses an additional tool to populate subtasks and database facts in Levels 2 and 3, respectively. Details on prompts and tools are provided in Appendix~\ref{appdx:prompts}. 

By abstracting domain and schema-specific insights from each interaction and storing them in memory, the agent can efficiently apply prior knowledge to new tasks, reduce redundancy, and improve its reasoning transparency. This not only enhances performance on future queries but also supports scalability and aligns with established human learning paradigms.

\subsubsection{Procedural Reasoning}

We evaluate two levels of procedural reasoning in our Learning Agent: no procedural reasoning (NP) and reasoning (P). In the NP condition, the agent operates under a templated prompt and is required to generate and refine SQL in a single step. The prompt includes the database schema and retrieved memories, with retrieval performed based on cosine similarity of the NLQ. In this setting, the retrieval query is fixed by the template, rather than being specified by the LLM. This setup is akin to a simple LLM baseline that uses in-context learning, and serves as a strong point of comparison.

In contrast, the P setting allows the agent to proceed step by step, with the specific sequence of actions (e.g., decomposition, planning, retrieval, or assembly) determined by the agent itself rather than fixed in advance. Here, the LLM specifies retrieval queries and the granularity for memory retrieval, giving it substantially more flexibility in accessing relevant memories without the need to tune a similarity metric. This procedural reasoning also introduces greater variability in the agent’s trajectory, since outcomes depend more heavily on the correctness of intermediate steps.

\begin{table*}[h!]
\centering
\resizebox{\textwidth}{!}{%
\begin{tabular}{c c c c c c c c c c c c}
\toprule
& \multicolumn{4}{c}{Memory} & \multicolumn{1}{c}{Reasoning}  & \multicolumn{5}{c}{Execution Accuracy}\\
\cmidrule(lr){2-5} \cmidrule(lr){6-6} \cmidrule(lr){7-11}
\makecell{Agent\\Label} 
  & \makecell{NLQ-\\SQL} 
  & \makecell{Distilled\\Knowledge}
  & Subtasks
  & \makecell{Database\\Facts}
  & \makecell{Procedural\\Reasoning}
  & \makecell{Initial\\Performance} 
  & \makecell{Online\\Performance} 
  & \makecell{Final\\Performance} 
  & $\Delta_i$ 
  & $\Delta_o$ \\
\midrule
NP-0 & \cmark & \xmark & \xmark& \xmark& \xmark& 35.1  & 92.0 & 77.4 &  42.3 &  -14.6 \\
NP-1 & \cmark & \cmark & \xmark &\xmark &\xmark & 35.1  & 92.0 & 64.5 &  29.4 &  -27.0 \\
P-0  & \cmark&\xmark &\xmark &\xmark & \cmark & 32.9  & 91.3 & 76.9 &  46.2 &  -13.4 \\
P-1  & \cmark & \cmark & \xmark & \xmark &\cmark & 32.9  & 91.3 & 79.1 &  48.9 &  -10.7 \\
P-2  & \cmark & \cmark & \cmark &\xmark &\cmark &  32.9 & 91.3 & 81.3 &  49.5 &  -10.1 \\
P-3  & \cmark& \cmark& \cmark& \cmark& \cmark &  32.9 & 91.3 & \textbf{81.7}  &  \textbf{48.8} &  \textbf{-9.6}  \\ 
\bottomrule
\end{tabular}
}
\caption{Execution accuracy when the same set of questions is used for each phase. $\Delta_i$ is the difference between the final and initial performance, while $\Delta_o$ denotes the difference between the final and online performance. Performances reported are averages of 3 runs.}\label{tab:same_q}
\end{table*}
% }

\subsection{Human Proxy Agent (HPA)}

To provide scalable, human-like feedback during SQL generation, we design a Human Proxy Agent (HPA) that emulates an expert database user. The HPA is built on an LLM, which we use GPT-4o, and is provided with the NLQ, ground-truth SQL query, annotated expert ``evidence'', and database schema. The HPA is then prompted to create a chain-of-thought solution to reference when providing feedback.

When the learning agent proposes a SQL query, the HPA executes it against the database and compares the output to the ground truth. If the query is correct, the HPA confirms completion, otherwise, it generates natural language feedback that identify errors in the logic of the SQL and prescribes corrections, simulating how a human expert would guide the agent. This feedback is sent to the learning agent, which refines its query accordingly. The process repeats until the correct SQL is generated, the conversation reaches 25 steps, or the maximum context length of the LLM is exceeded.

\section{Empirical Results}

We first describe our experimental setup and evaluation criteria in Section~\ref{sec:eval}. In Section~\ref{sec:ablations}, we analyze the effects of increasing memory complexity, procedural reasoning, and their interaction. Sections~\ref{sec:sameq} and~\ref{sec:newq} evaluate the agent’s online and final performance on the same and new questions, respectively, highlighting its ability to retain feedback and generalize across tasks within a domain. Section~\ref{sec:learning_curves} examines how performance scales with the number of online instances, and Section~\ref{sec:failure} analyzes error types and prevalence to understand what the agent is learning.

\subsection{Evaluation} \label{sec:eval}

We evaluate the Learning Agent on the BIRD Dev set \citep{li2024can}, which contains 11 databases spanning a variety of domains. We focus on BIRD rather than other benchmarks because it provides annotated expert evidence for each question, which is crucial for the HPA to generate realistic feedback containing tacit knowledge.

For each database, we randomly divide the questions into two equal splits—\emph{train} and \emph{test}—with $T$ samples each. Our evaluation proceeds in three phases, as illustrated in Figure~\ref{fig:test_setup}: initial performance (before learning), online performance (refinement through human feedback), and final performance (after learning).

\paragraph{Initial Performance.} We run the offline setup with an empty memory store, establishing the performance based on database schema understanding and pretrained LLM knowledge. 
\paragraph{Online Performance.} We run the online setup. The Learning Agent receives natural language feedback from the HPA until the generated SQL is correct or a maximum number of iterations is reached. Upon successful generation, the instance is saved to the agent memory store for future use, and if available, the Learning Agent distills the feedback as knowledge.
\paragraph{Final Performance.} We run the offline setup with the memory store built during the \textit{Online Performance} phase. This final performance demonstrates the performance based on the database schema understanding, pretrained LLM knowledge, and the retrieved knowledge from the memory store.

The difference in the \textit{Final Performance}, $\bar{Z}_{T}$, and the \textit{Initial Performance}, $\bar{Z}_0$ can be attributed to the usage of the memory store. We also evaluate $\bar{Z}_t$  for $t \in \{1, \dots, T\}$ to quantify the impact of the quantity of instances used to construct the memory store.

\subsection{Memory Retention and Generalization}\label{sec:ablations}

\begin{figure*}
    \centering
    \includegraphics[width=\linewidth]{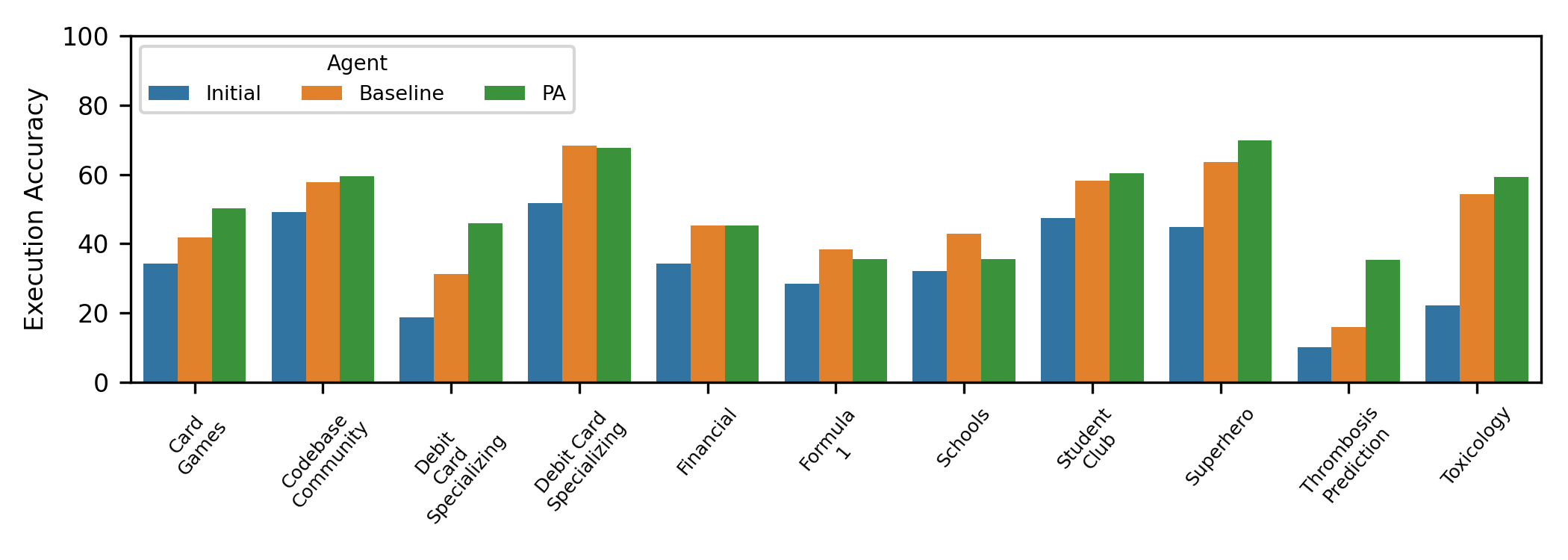}
    \caption{Execution accuracy by database for the Initial (NP-0, no memory), Baseline (NP-0), and Procedural Agent (PA, Agent label P-3).}
    \label{fig:placeholder}
\end{figure*}

We evaluate the Learning Agent’s continual learning capabilities in two complementary settings. First, in Section~\ref{sec:sameq}, we assess retention on previously seen questions, measuring how effectively the agent retrieves and leverages past experiences to reproduce correct SQL queries. Second, in Section~\ref{sec:newq}, we evaluate generalization to new questions, examining the agent’s ability to reuse acquired knowledge on unseen tasks within the same domain. 

\subsubsection{Retention on Previously Seen Questions} \label{sec:sameq}

Before considering generalization to new tasks, it is imperative that the Learning Agent can successfully complete a task that it has previously completed successfully. To this end, we evaluate the initial, online and the final performance on the train split for each database. Note that in the online setting, the HPA is providing direct feedback for refinement, and will continue to provide feedback until the Learning Agent generates the correct SQL query. During the offline phase, to determine final performance, the Learning Agent must effectively retrieve and reason over its stored memories, accurately identifying and leveraging relevant prior experiences even when presented alongside other, potentially distracting experiences.

Table~\ref{tab:same_q} shows results for this setting. Both the NP and P agents performance similarly for the initial and online performances, demonstrating similar capabilities in the absence of any memories. The procedural reasoning enables a superior ability for the Learning Agent to effectively utilize memories on the same tasks. This is especially apparent in the agent trajectories, where P-type agents can explicitly reason about the retrieved memories and recognize that the current question matches that of the retrieved memory. We note, that a considerable performance degradation occurs for NP-1 vs NP-0, as compared to any of P-1, P-2, P-3, to P-0. This suggests a positive interaction effect between knowledge distillation and the procedural reasoning of the Learning Agent. The addition of the subtask and database fact memories stores further improves the final performance. 

\subsubsection{Generalization to New Questions}\label{sec:newq}

We next consider the setting where the question splits differ between the online and final performance phase. We perform the initial, and final phases on the test split, and the online phase on the train split. This setting evaluates the Learning Agent's ability to generalize previously acquired experiences to new tasks in the same domain. This scenario is more representative of practical use, where the agent faces new questions sequentially.

\begin{table}[h!]
\centering
% \resizebox{\textwidth}{!}{%
\begin{tabular}{c c c c c c c c}
\toprule
\makecell{Agent\\Label} & \makecell{Initial\\Performance}  & \makecell{Final\\Performance} & $\Delta_i$ \\
\midrule
NP-0 & 34.1  & 47.0 &  12.9 \\
NP-1 & 34.1  & 43.7 &  9.6 \\
P-0  & 32.4  & 49.4 &   17.0 \\
P-1  & 32.4  & 50.8 &  18.4 \\
P-2  & 32.4  & 49.5  &  17.1 \\
P-3  & 32.4  & \textbf{51.3} & \textbf{18.9}\\ 
\bottomrule
\end{tabular}

\caption{Execution accuracy by Agent Label when questions differ between the online and offline phases, averaged across 3 runs.}\label{tab:new_q}
\end{table}
% }

All agent settings show an increase in final performance relative to the initial performance. Procedural reasoning alone improves accuracy by 2.4\%, underscoring its role in continual learning. In contrast, adding distilled knowledge without reasoning is detrimental to performance, as observed in the difference between NP-1 and NP-0. However, when combined with procedural reasoning, distilled knowledge yields a substantial performance gain, indicating a positive interaction effect. The inclusion of additional memory granularities further enhances performance, with the P-3 setting, which we henceforth refer to as the \emph{Procedural Agent} (PA), achieving the highest accuracy on both previously seen and new tasks. These results demonstrate that procedural reasoning, together with a multi-level memory store, enables the most effective continual learning.

\begin{figure*}
    \centering
    \includegraphics[width=\textwidth]{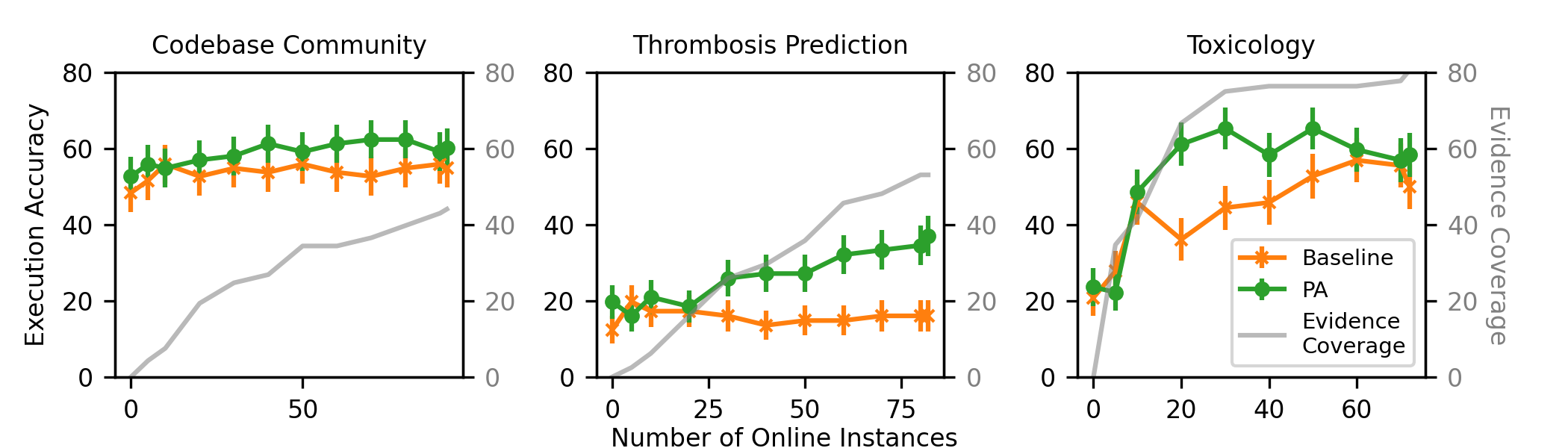}
    \caption{Execution accuracy as a function of the number of online instances used to construct the memory store. ``Baseline'' corresponds to the non-procedural agent (NP-0), and ``PA'' corresponds to the full Procedural Agent (P-3). Error bars indicate $\pm$1 standard deviation. ``Evidence Coverage'' denotes the proportion of test questions that have at least one corresponding question in the memory store whose annotated evidence field has cosine similarity $\geq$ 0.9 to that of the test question.}
    \label{fig:learning_curves}
\end{figure*}

\subsection{Impact of Memory Size on Learning}\label{sec:learning_curves}

Next, we examine how the number of instances in the memory store impacts performance. Figure~\ref{fig:learning_curves} shows execution accuracy as a function of the number of instances used to construct the memory store across three databases. In more cases than not, the non-procedural baseline exhibits faster initial learning than the procedural agent, particularly within the first 10–20 instances. This likely reflects the fact that the procedural method requires more information to populate knowledge across three levels of granularity, rather than a single level. However, as the number of instances increases, the procedural method ultimately surpasses both non-procedural settings. Future work may investigate hybrid approaches that combine the rapid early gains of the non-procedural agent with the long-term advantages of procedural reasoning, enabling strong initial performance while progressively leveraging multi-level memory representations. 

When observing the overlap in tacit knowledge between the instances used to populate the memory store and the test questions, measured as ``evidence coverage'' in Figure~\ref{fig:learning_curves}, the  execution accuracy of the PA correlates  with this metric, suggesting that the PA does effectively capture tacit knowledge. As evidence coverage plateaus or increases at a slower rate, the incremental benefit of additional tacit knowledge is outweighed by the increased complexity of retrieving from a larger memory store. This suggests that effective management of the memory size is an important direction for future work. Similar plots for all databases are provided in Appendix~\ref{appdx:addtl_results}.

\subsection{Failure Analysis}\label{sec:failure}

\begin{figure*}
    \centering
    \includegraphics[width = \textwidth]{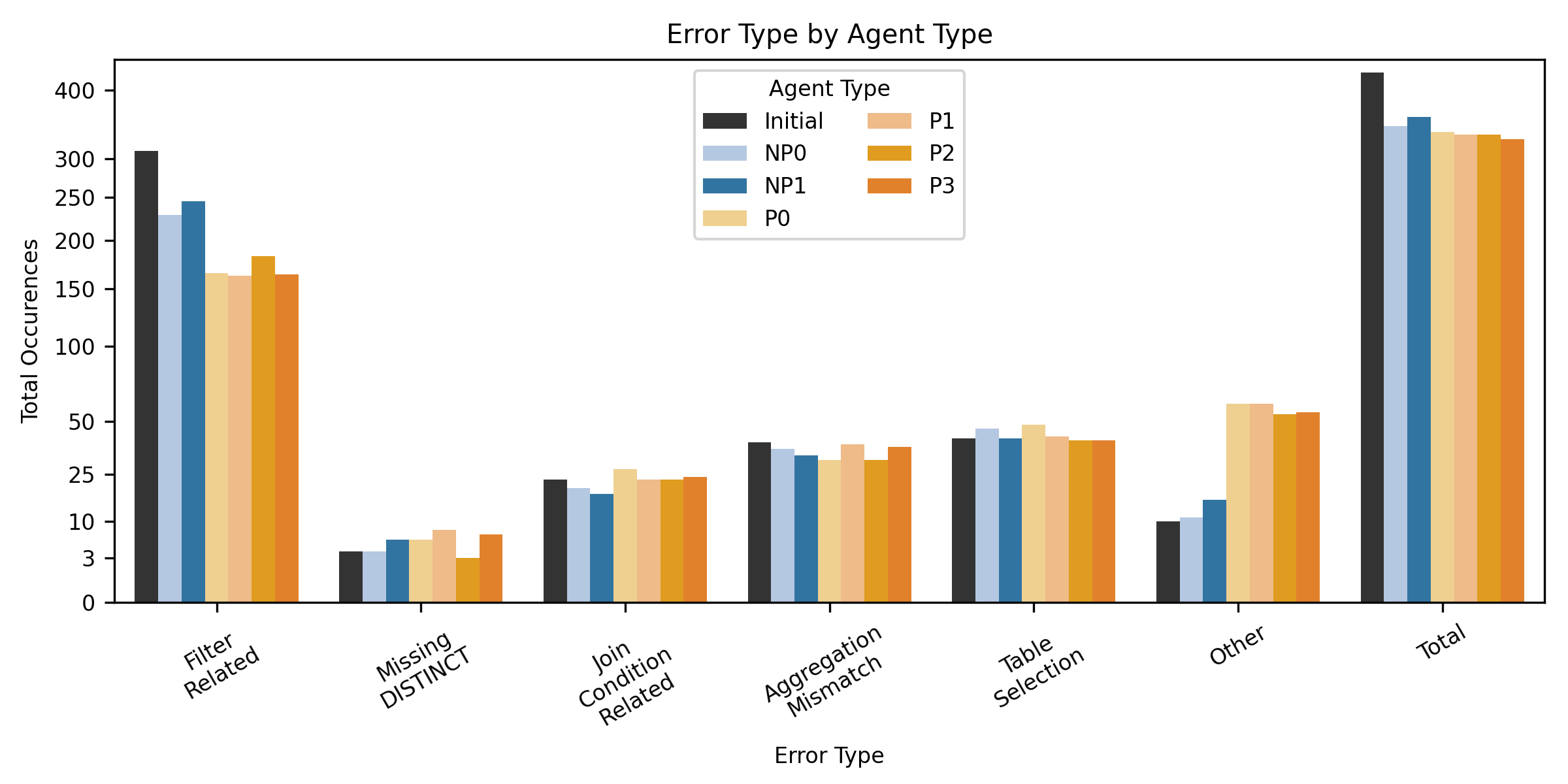}
    \caption{Breakdown of error types for each agent configuration. Bars show the total counts of common error categories across different agents.}
    \label{fig:error_types}
\end{figure*}

To better understand the sources of performance gains, we analyze the primary failure types across different agents, focusing on which errors are most effectively resolved. Figure~\ref{fig:error_types} illustrates the distribution of five common error categories, plus an “Other” category. When comparing agents that exclude versus include distilled knowledge in memory, we find that excluding distilled knowledge leads to more effective resolution of syntax-level errors, such as issues with filters or the use of the DISTINCT clause. Conversely, including distilled knowledge enhances the agent’s ability to correct errors related to reasoning and schema comprehension, such as join conditions and aggregations. This indicates that while distilled knowledge—introduced via prompt tokens—may occasionally distract from detailed SQL syntax, it can explicitly surface reasoning patterns that are not always apparent from NLQ-SQL pairs alone.

The addition of multi-level granularity in memory and procedural reasoning enables the agent to balance these approaches, reducing both syntax and reasoning related errors. Although the error rate for DISTINCT clause usage increases slightly, this typically occurs in queries that are otherwise syntactically and structurally correct and is outweighed by reductions in more critical errors, such as incorrect joins or table selection. This may also explain the lower prevalence of such errors in non-procedural methods. The rise in “Other” errors for agents employing procedural reasoning is primarily due to the selection of additional columns, which results in execution mismatches.

\section{Conclusions}

This work presents a novel framework for addressing the critical challenge of tacit knowledge acquisition in the text-to-SQL setting, focusing on a practical learning scenario. Our approach enables a Learning Agent to begin interacting with a human expert with minimal setup, iteratively generating and refining SQL queries through natural language feedback. By integrating a multi-level memory with procedural reasoning and guided knowledge distillation process, the agent can systematically store, retrieve, and reason over this knowledge, improving both recall of previously solved queries and generalization to new tasks within the same domain.

Empirical evaluation on the BIRD Dev set demonstrates that procedural reasoning significantly enhances memory utilization, while distilled knowledge focuses the agent on reasoning related errors that are difficult to capture through NLQ-SQL pairs alone. Multi-level memory granularities further improve performance by allowing flexible access to subtasks and database facts, enabling the agent to reconcile local query details with higher-level reasoning patterns. Collectively, these results highlight a practical, effective framework for learning in text-to-SQL environments. Beyond the text-to-SQL domain, this framework offers a blueprint for continual learning in structured reasoning tasks where expert feedback is available.

\section{Limitations}

While our results demonstrate the potential of this methodology, further research in the text-to-SQL domain is needed to fully assess its effectiveness on more complex databases and queries, especially in scenarios involving noisy feedback. Additionally, the dynamic nature of database schemas presents a challenge, as memories based on outdated schema knowledge may become less reliable over time. Our evaluation is also limited by its reliance on Execution Output, which does not always account for variations such as column ordering or the presence of unique versus duplicate values in responses.

Beyond text-to-SQL, we believe this methodology holds promise for a wide range of reasoning tasks that require domain-specific knowledge not accessible during LLM pretraining, such as general tool usage, mathematical reasoning, and code generation. Understanding the scalability of this approach is also essential, particularly as memory stores grow in size. We leave the design of memory management to future work.

Ultimately, we envision that this framework will enable the development of more adaptive, resilient, and human-informed AI systems, with significant potential to impact a variety of knowledge-intensive applications.

\section{Ethical Considerations}

\subsection{Ethical Impact}
In expanding autonomous agent capabilities, we raise some important ethical and societal considerations. Although there is still significant room for improvement, as robust autonomous systems that continuously learn in cold-start regimes could enhance accessibility to SQL generation and, we hypothesize, other logical tasks, it is crucial to proactively asses the broader economic and social implications (especially regarding employment) before deploying such technologies in real-world settings.

\subsection{Bias and Safety}
To ensure that autonomous agents do not inadvertently exclude or disadvantage any group, additional research is required to understand and address potential harmful behaviors. Though our contributions revolve mainly around memory design and task decomposition, comprehensive analysis is needed to detect and mitigate unintended biases in deployed systems. As of now, we do not explicitly test for this, but encourage adapters of our methodology to do so.

\subsection{Intended Uses}
We make use of BIRD, which serves as a research benchmark for evaluating various memory designs and procedural reasoning to understand whether an agent can continually learn in the text-to-SQL space. The models and methods presented in this paper are research prototypes and should be thoroughly tested and evaluated if applied to practical applications, especially in high-risk domains.

\section*{Disclaimer}
This paper was prepared for informational purposes by the Artificial Intelligence Research group of JPMorgan Chase \& Co. and its affiliates (``JP Morgan'') and is not a product of the Research Department of JP Morgan. JP Morgan makes no representation and warranty whatsoever and disclaims all liability, for the completeness, accuracy or reliability of the information contained herein. This document is not intended as investment research or investment advice, or a recommendation, offer or solicitation for the purchase or sale of any security, financial instrument, financial product or service, or to be used in any way for evaluating the merits of participating in any transaction, and shall not constitute a solicitation under any jurisdiction or to any person, if such solicitation under such jurisdiction or to such person would be unlawful.

\textcopyright  2025 JPMorgan Chase \& Co. All rights reserved

\section*{Acknowledgments}
We thank Leo Ardon for valuable input and suggestions on this paper.

\bibliography{references}

\appendix
\section{Required Implementation Details}\label{appdx:licenses}
\subsection{Datasets and Models}

\begin{table}[H]
    \centering
    \small
    \begin{tabularx}{\columnwidth}{|c|X|X|}
    \hline
        Entity & Name & License \\
    \hline
        Dataset & BIRD \cite{li2024can} & CC BY-SA 4.0 \\
        \hline
        Model & gpt-4o-2024-05-13 \cite{openai2024gpt4o} & Enterprise license \\
        \hline
        Model & text-embedding-ada-002 \cite{openai2022ada002} & Enterprise license \\
    \hline
    \end{tabularx}
    \caption{Datasets, Models and licenses}
    \label{tab:licenses}
\end{table}

\subsection{Computational Budget, packages, versions used}\label{appdx:implementation}

\subsubsection{Computational Budget}
Experimentation used a 16GB CPU as the LLM, GPT-4o was accessed via API. Total run time was approximately 6 CPU days. Total API costs for develop and reported experiments were approximately $\$7000-\$8000$.

\subsubsection{Packages}
For our multi-agent design, we make use of a closed-source package, based on LangChain. Although the specific package is not open-sourced, we provide highly detailed prompts to facilitate replication. We use Python 3.11.9 with the following packages

\begin{table}[H]
    \centering
    \small
    \begin{tabular}{|c c|}
    \hline
        Package & Version \\
    \hline
        openai & 1.81.0 \\
        pandas & 2.2.2\\
        numpy & 1.26.4\\
        json5 & 0.9.14\\
        langchain & 0.3.25\\
        langchain-community & 0.3.24\\
        llama-index & 0.12.31\\  
    \hline
    \end{tabular}
    \caption{Python Packages and Versions Used}
    \label{tab:packages}
\end{table}

\section{Additional Agent Details}\label{appdx:prompts}
\subsection{Non-procedural agent with Level 0 Memory (NLQ-SQL pairs)}\label{appdx:base_agent}

% Define a simple prompt box style
\tcbset{
    promptstyle/.style={
        colback=gray!10, % Light gray background
        colframe=black, % Black frame
        sharp corners,
        boxrule=0.5pt, % Thin border
        fonttitle=\bfseries,
        width=\textwidth,
        left=0pt,
        right=0pt,
        top=2pt,
        bottom=2pt,
        boxsep=5pt,
        breakable
    }
}

The agent is an LLM-based agent that has two tools that it can call. The \texttt{generate\_sql} tool takes as input the natural language question (NLQ) from the user and uses a templated prompt to perform an LLM call. This template consists of at most three similar NLQ-SQL pairs (as defined by the cosine similarity between the vector embeddings of the current NLQ and the example NLQs) from a set of saved examples from previous, successful runs if they exist in the agent's historical memory, and the database schema, and asks the LLM to generate a SQL query. The agent is specifically instructed to generate SQL queries in adherence to the database schema. 

If the agent receives feedback from the human expert, then it will call its \texttt{refine\_sql} tool. This tool follows a similar template, but instead seeks to refine the generated SQL based on natural language feedback, rather than generate one from scratch. The human expert may continue to provide feedback until the agent generates the correct SQL. The memory store, $M_t$, consists of a vector database of NLQ-SQL pairs that the agent was successful in generating, enabling continual learning for future queries. The index is the NLQ, which enables retrieval by the semantic similarity of the example NLQs and the current NLQ.

\paragraph{Agent System Prompt, Offline}
\begin{tcolorbox}[breakable]
You are an agent who generates SQL queries for a given natural language questions to query a given set of tables. Here are some tips for generating the SQL queries successfully.\\
- Each SQL query needs to be generated only once.\\
- Only share the final SQL query which is the SQL query for the last sub question with the human user.\\
- Be sure to refine the SQL after generation to resolve any errors that you can fix yourself. You can use the refine sql tool and pass in a blank string for the feedback argument. \\
- Do not change the original question provided to you when providing the question in tool calls.\\
- Do not use the generate SQL tool if you have already generated a query. You must refine the sql at that point.\\
- Use the human\_return communcation spec when sending a message to the human. Please separate your message and the generated SQL query accordingly.
\end{tcolorbox}
\paragraph{Agent System Prompt, Online}
\begin{tcolorbox}[breakable]
You are an agent who generates SQL queries for a given natural language questions to query a given set of tables. Here are some tips for generating the SQL queries successfully.\\

- Start by using generating a SQL query, using the generate\_sql tool call. Do not return control to the human after this. \\
- Then request feedback from the Expert agent so that you can to receive feedback. You must request feedback from the Expert agent at least once before returning the query to the human. \\
- You must refine the SQL query based on the feedback provided by the expert agent. Use the refine\_sql tool for this.\\ 
- You may request feedback from the expert multiple times.\\
- You MUST get confirmation from the Expert that the generated SQL query is correct before returning it to the human. Never return and unconfirmed SQL query!!!!\\
- Do not change the original question provided to you when providing the question in tool calls.\\
- Do not use the generate SQL tool if you have already generated a query. You must refine the sql at that point.\\
- Do not ask the expert agent to generate the SQL query or ask the ground truth query.\\
- Use the return\_gen\_sql Communication Spec when communicating with the Expert agent.\\
- Use the human\_return communcation spec when sending a message to the human. Please separate your message and the generated SQL query accordingly.
\end{tcolorbox}
\paragraph{Generate SQL Prompt}
\begin{tcolorbox}[breakable]
You are a SQL query generation agent. Instructions for your task are as follows:\\

Generate a single SQL query which answers the following question that can be executed in a SQLITE3 database. \\

--------------------------------\\
Examples\\

\{examples\}\\

--------------------------------\\

\#\# Question:\\
\{question\}\\

\#\# Database schema and information:\\
\{database\_schema\}\\

\#\# Task Output:\\

SQL Query:

\end{tcolorbox}
\paragraph{Refine SQL Prompt}

\begin{tcolorbox}[breakable]
You are a SQL query generation agent with the capability to refine SQL queries given feedback. Instructions for your task are as follows: \\

Refine the given OLD SQL for the following QUESTION based on the FEEDBACK provided. \\
- If a SQLite error is provided, you must take that error into account and refine the SQL based on the error as well.\\
- Only produce a single SQL query.\\

--------------------------------\\
Example\\

\#\# Question:\\
What is the eligible free rate of the 10th and 11th schools with the highest enrollment for students in grades 1 through 12?\\

\#\# Old SQL\\
SELECT CAST(`Free Meal Count (K-12)` AS REAL) / `Enrollment (K-12)` FROM schools ORDER BY `Enrollment (K-12)` DESC LIMIT 9, 2\\

\#\# Feedback\\
You should select from the table 'frpm', not schools\\

\#\# Refined SQL\\
SELECT CAST(`Free Meal Count (K-12)` AS REAL) / `Enrollment (K-12)` FROM frpm ORDER BY `Enrollment (K-12)` DESC LIMIT 9, 2\\

--------------------------------\\
\#\# Question \\
\{question\}\\
\#\# Database name \\
\{database\_name\}\\
\#\# Database schema\\
\{database\_schema\}\\
\#\# Examples\\
\{examples\}\\
\#\# Feedback \\
\{feedback\}\\
\#\# old SQL\\ 
\{generated\_sql\}\\

Now please fixup old SQL and generate new SQL again. \\

\#\# Task Output:\\

\end{tcolorbox}

\subsection{Procedural Agent}

The \emph{Procedural Agent} (PA) differs significantly in that it does not use a fixed workflow for SQL generation and refinement. Instead, the agent is allowed to reason step by step, and make use of its \texttt{find\_memory} and \texttt{save\_memory} tools, interweaving retrieval, decomposition, and reasoning steps on the fly in a procedural manner. Since the agent has more autonomy in constructing a SQL query, we correspondingly expand the memory to three vector databases, containing similar questions, subtasks, and database facts, respectively. The question memory store corresponds a store of NLQ-SQL pairs that are enhanced with \emph{distilled knowledge}. This enhancement consists of tacit knowledge the agent extracts at the conclusion of an interaction with the human and is based on the conversation with the human, the database schema, and the correct final generated SQL.The subtask memory store contains task description-SQL chunk pairs, which correspond to elements of a decomposed SQL query, inspired by the notion of example coverage explored by \cite{gupta2023coveragebased}. Lastly, the database-fact memory store contains description-hint pairs, ranging from schema linking, table contents, filter values and SQL operation hints. For constructing entries to the subtask and database-fact memory stores, the agent makes use of the \texttt{save\_memory} tool, and must specify the index, the value, and type of memory being stored.

 The PA starts by retrieving similar questions and reasoning about the utility of the examples retrieved (database schema and general SQL reasoning instructions are specified in the agent's system prompt). It then decomposes the SQL query into steps, and performs retrieval on a \textit{similar subtask} memory store, grounding components in previous experiences. Database schema and general SQL reasoning instructions are specified in the agent's system prompt.During this step, the PA may also retrieve memories on a \textit{database-fact} memory store in order to fill in knowledge gaps that may not be tied directly to a question or subtask. It then assembles the SQL query and verifies that it does not violate the database schema, as with the non-procedural Agent with Memory Level 0 (NP-0).

\paragraph{Agent System Prompt Offline}
\begin{tcolorbox}[breakable]
Your name is "Learning SQL Agent". You are an agent who generates SQL queries for a given natural language questions to query a given set of tables. Here are some tips for generating the SQL queries successfully.\\
You will have access to the natural language question, the database schema, and some knowledge stores. Your generated SQL must adhere to the database schema. The query should be executable on a SQLite database.\\

YOU MUST ADHERE TO THE FOLLOWING STEPS, IF YOU DEVIATE FROM THESE STEPS YOU WILL FAIL.\\
1. You should first find the most similar questions. Use the find memory\_tool for this step. Evaluate how similar the memories are and elaborate on what information may be explicitly useful.\\
2. You should then decompose the SQL query down step by step. This should create a logical plan of how the SQL query will come together.\\
3. Based on this plan, you should use find\_memory tool to find related sub\_tasks and database\_facts. You must use this tool before generating the SQL. It has useful information you have accumulated over time.\\
4. After these steps, you should reason about generating SQL query. Think out loud, step by step. Feel free to go back to searching for more memories to work on the query. YOU SHOULD REALLY USE THE MEMORIES.\\
5. Generate the SQL query. This will be a candidate answer, contained in a thought, that you will reason about before sharing.\\
6. Although you are breaking the SQL query down step by step and creating queries for each step, you must avoid using subqueries in the final SQL. You should use joins when possible. Assemble these subtasks through joins. Rework the query if necessary.\\
6. Verify the generated SQL query by referencing memories. Elaborate on if any portion of the SQL query conflicts information from memories, and make the necessary updates.\\
7. Once you are satisfied with the SQL query, share the SQL query with the human, along with an explanation of the logic of the SQL query.\\

Some helpful tips:\\
- Do not generate and immediately share it with the human. You are REQUIRED pose a candidate answer to yourself, and check it first!\\
- You can reason with yourself by sending a message to yourself. Note that if the most recent message is from yourself, "Learning SQL Agent", you should build off of that message, not just repeat it.\\
- Use the human\_return communication spec when sending a message to the human. Please separate your message and the generated SQL query accordingly.\\
- Be especially careful about the values used for filtering.\\
- Do not try to assemble a message and thought at the same time. Do these actions one at a time.\\

Here is the database schema:\\
\{db\_schema\}
\end{tcolorbox}
\paragraph{Agent System Prompt Online}
\begin{tcolorbox}[breakable]
Your name is "Learning SQL Agent". You are an agent who generates SQL queries for a given natural language questions to query a given set of tables. Here are some tips for generating the SQL queries successfully.\\

- You will have access to the natural language question, and the database schema. Your generated SQL must adhere to the database schema. The query should be executable on a SQLite database.\\
- You should first break the SQL query down step by step. \\
- You should reason about a plan to assemble the SQL query by completing these steps. \\
- Although you are breaking the SQL query down step by step and creating queries for each step, you must avoid using subqueries in the final SQL. You should use joins when possible. Assemble these subtasks through joins.\\
- Be especially careful about the values used for filtering.\\
- You can message a database expert agent to request feedback. You should get confirmation that the query is correct before returning it to the human. You MUST message the expert at least once.\\
- When the expert provides feedback, you should reflect on the feedback and plan a revision. With this plan, refine the SQL query to incorporate the feedback. You must request feedback after any refinement.\\
- Once the expert says the sql query is correct, only then can you share the SQL with the human. Share an explanation of the SQL query as well.\\
- You can reason with yourself by sending a message to yourself. Note that if the most recent message is from yourself, "Learning SQL Agent", you should build off of that message, not just repeat it.\\
- Use the human\_return communication spec when sending a message to the human. Please separate your message and the generated SQL query accordingly.\\
- Do not save memories until the human instructs you to.\\

- Do not try to generate a message and thought at the same step. YOU CAN ONLY DO ONE OF THESE PER STEP!!!!\\

Here is the database schema:\\
\{db\_schema\}
\end{tcolorbox}

\paragraph{Distilling Knowledge}

Upon successful generation of the SQL query, one of the following messages is sent to the learning Agent. If the learning agent did not receive any feedback (it generated a correct SQL initially), the following message is sent:
\begin{tcolorbox}[breakable]
Please summarize what you learned from generating this SQL query. Think about it out loud in many steps, and then give me a concise and specific summary. \\

Do these steps sequentially. The output of the previous step is used for the subsequent steps.\\
                        
1. First send yourself a message which describes any domain specific knowledge that was needed for generating this SQL query. Explicitly highlight what reasoning steps that you took that were correct.\\
2. Second, send yourself a message which details what database knowledge was revealed. Link this is to the schema. Be very specific, and SQL heavy.\\
3. Third, look at the thoughts generated by steps 1 and 2. Rephrase the findings as facts that can be useful in the future.\\
                        
The only response that I want back is the set of facts. Leave the "generated\_sql" field empty in your response.
\end{tcolorbox}

Otherwise, the following message is sent:
\begin{tcolorbox}[breakable]
Please summarize what you learned from interacting with the Expert agent. Think about it out loud in many steps, and then give me a concise and specific summary. \\

Do these steps sequentially. The output of the previous step is used for the subsequent steps.\\
                        
1. First send yourself a message which describes the mistakes you made. This should explicitly state what component of the SQL query was incorrect. You should also pinpoint where this error occurred in your reasoning.\\
2. Second, send yourself a message which details what database knowledge was revealed through feedback to correct this mistake. Link this is to the schema. Be very specific, and SQL heavy. This may contain information about the reasoning you had about the database.\\
3. Third, look at the thought generated by step 2. Rephrase the findings as facts. Provide some context as to when these facts would be useful in the future.\\
                        
The only response that I want back is the set of facts. Leave the "generated\_sql" field empty in your response.
\end{tcolorbox}

The learning agent then responds with distilled knowledge. This knowledge is added to the NLQ-SQL pair and is saved to the similar\_question memory store. Next, the following message is sent to the learning agent to save memories to the sub task and database fact memory stores:

\begin{tcolorbox}[breakable]
Now save the distilled knowledge, one chunk at a time, for future use by using your save\_memory tool. You can decompose the SQL query and save the chunks as sub tasks as useful reusable components in the future. You can also save database\_facts, which contain useful information that may be necessary in the future in a more general set of tasks. \\

You can save up to 5 memories. Plan and decide which are most important before you start. Here are some helpful tips:\\
                    - Be sure to make the index of the memory useful. A useful index is one that will be easy to search for in the future. For example, dont just say "join table1 and table2". Say something like "join table1 and table2 by attribute", where attribute is the common column.\\
                    - When saving a similar subtask, the index should generally describe the subtask. Always make the passage contain a chunk of SQL. The index must contain information about the values filtered for in the SQL if you keep the values in the passage. Otherwise, you should replace the filters with placeholders, and the index should reflect the generality.\\
                    - Don't save basic SQL knowledge like "how to limit". This is information you already know. You could instead say "how to sort for \_\_column name\_\_" and then the passage can show that you order and use a desc and limit clause in a SQL query.\\

                    Some examples of what not to do:\\
                    
                    'query\_string': 'use of JOIN operations to connect tables',\\
                    'knowledge\_string': 'JOIN operations are used to connect tables based on common columns to retrieve related data.',\\
                    'memory\_type': 'database\_fact'\\

                    Reason: This does not provide any useful information about the database and is just common SQL knowledge.\\

                    'query\_string': 'filter by exact text values in SQL',\\
                    'knowledge\_string': "When constructing SQL queries involving the 'frequency' column, it is essential to use the exact text values defined in the schema for accurate filtering. This ensures the query retrieves the intended results based on the issuance frequency.",\\
                    'memory\_type': 'database\_fact'\\

                    Reason: You should instead save "filter by issuance frequency" as the index, and the the passage should be showing the portion of the SQL query of how to do this. If available, you should specify what the filter value corresponds to in the index.\\

                    'query\_string': "filter accounts by frequency 'POPLATEK TYDNE'",\\
                    'knowledge\_string': "SELECT * FROM account WHERE frequency = 'POPLATEK TYDNE';",\\
                    'memory\_type': 'similar\_subtask'\\
                    
                    Reason: The query string is not helpful since you would need to know to search for POPLATEK TYDNE in the first place. The tricky part is knowing what this value semantically means. Instead the query string should be "filter accounts for weekly issuance"
                    
\end{tcolorbox}

\onecolumn
\section{Memory Tools}\label{appdx:memory_tools}
The save\_memory tool is used at the conclusion of an online iteration, after the PA has distilled the knowledge. Note that this tool is used for sub\_task and database\_fact memory types, as the similar\_question memory type is hardcoded to prevent question rewording and to provide clarity between questions and subtasks. \newline

\newtcblisting{pythontools}[1][]{
  colback=blue!5,
  colframe=blue!75,
  sharp corners,
  boxrule=0.5pt,
  listing only,
  listing engine=listings,
  listing options={
    language=Python,
    basicstyle=\ttfamily\small,
    breaklines=true,
    showspaces=false,
    keywordstyle=\color{blue},
    commentstyle=\color{gray},
    stringstyle=\color{red},
    frame=none,
    numbers=none,
    keepspaces=true,
    columns=fullflexible,
    % fixes underscores and triple quotes
    literate=
      {\\_}{{\textunderscore}}1
      {_}{{\textunderscore}}1
      {'''}{{\textquotesingle\textquotesingle\textquotesingle}}1
      {"}{{{\color{red}"}}}1,
  },
  title=\textbf{Python Tool Definition},
  enhanced,
  drop shadow southeast,
  #1
}

\begin{pythontools}[title={Save Memory Tool}]

def save_memory(self, query_string:str, knowledge_string: str, memory_type : str) -> str: 
'''
Tool for saving information from the current iteration for future in-context learning. When using this tool, do not use any aliases.
Args:
- query_string (str): the piece of text you will be used as an index in a vector db. This is NOT the sql query. For example, "how to calculate eligible free rate".
- knowledge_string (): The knowledge attached to the index in the vector database. This can be a SQL query, a snippet of a SQL query, or a database fact.
- memory_type (str): One of ['similar_subtask', 'database_fact'], Specifies what the type of information you are searching for.
'similar_subtask' memory type is helpful for understanding how to perform particular operations on this database. This information is critical when assembling a SQL query.
'database_fact' memory type is helpful for understand critical database information which you may have over looked in the past. This is incredibly useful for ensuring filters are correct.
'''
    if self.memory_counter >=5:
        return "You have exceeded the number of memories that you can save for this question. Do not save any more memories."
    memory_store = load_vector_db(memory_type)
    if query_string not in memory_store.index:
        memory_store.insert(query_string, knowledge_string)
        memory_store.save()
        self.memory_counter += 1
    return "Saved."
\end{pythontools}

The find\_memory tool is used to retrieve memories during offline testing.
\begin{pythontools}[title={Find Memory Tool}]
def find_memory(self, query_string:str, memory_type : str) -> str: 
    '''
    Tool for retrieving information from previous experiences for in-context learning.
    
    Args:
    - query_string (str): the piece of text you want to search for. This is NOT the sql query. For example, "how to calculate eligible free rate".
    - memory_type (str): One of ['similar_question', 'similar_subtask', 'database_fact'], Specifies what the type of information you are searching for.
    
    'similar_question' memory type is helpful for seeing natural language question-sql query pairs. This is helpful for planning. DO NOT CHANGE THE WORDING OF THE NATURAL LANGUAGE QUESTION FROM HOW THE HUMAN STATED IT. DO NOT PARAPHRASE.
    'similar_subtask' memory type is helpful for understanding how to perform particular operations on this database. This information is critical when assembling a SQL query.
    'database_fact' memory type is helpful for understand critical database information which you may have over looked in the past. This is incredibly useful for ensuring filters are correct.
    '''
    memory_store = memory_store.load(memory_type)

    return memory_store.retrieve(query_string, k = 3, max_distance = 0.28)
\end{pythontools}

\twocolumn
\newpage
\onecolumn
\section{Additional Results}\label{appdx:addtl_results}

\begin{figure}[H]
    \centering
    \includegraphics[width=0.7\linewidth]{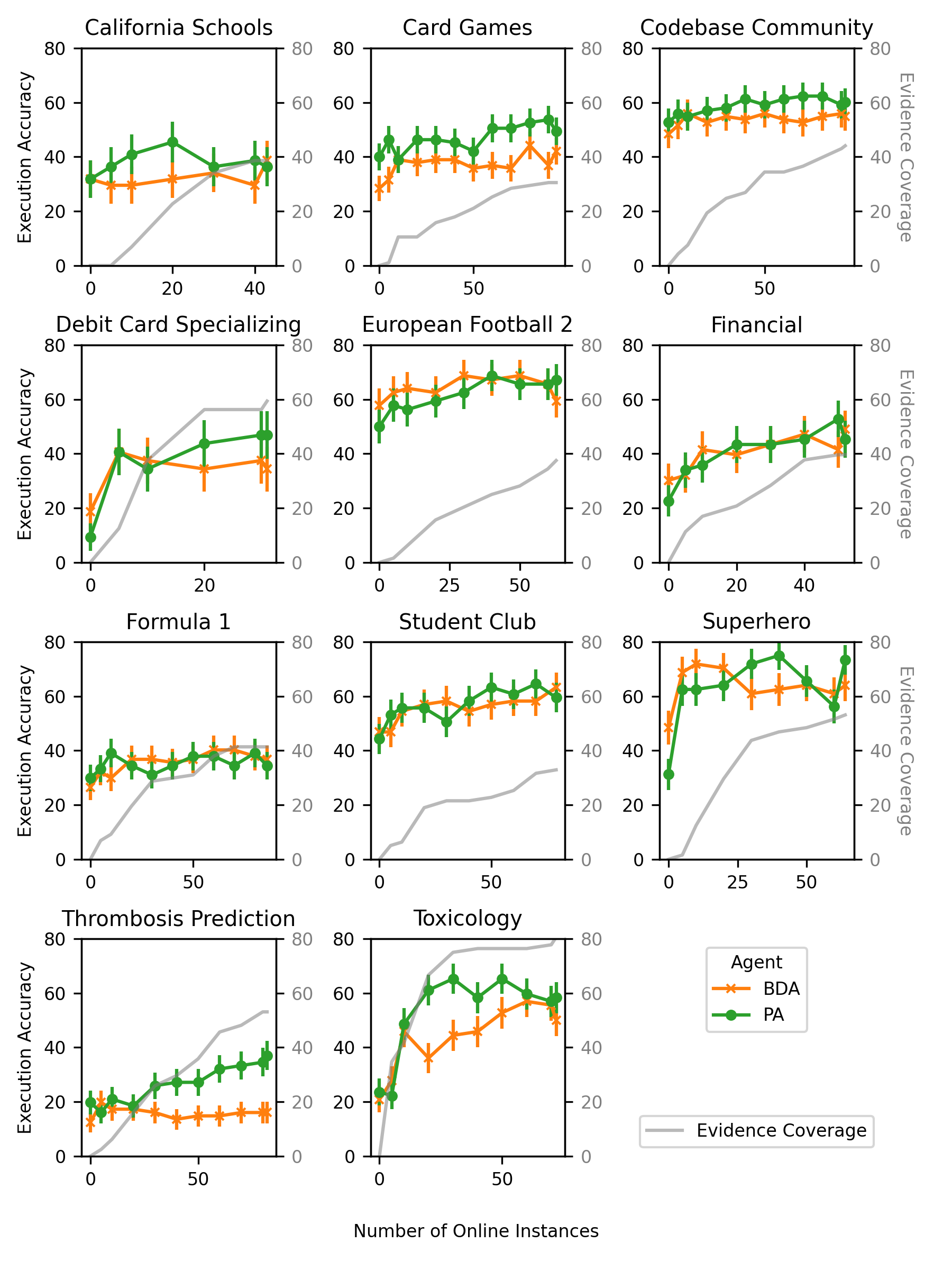}
    \caption{Execution accuracy as a function of the number of online instances used to construct the memory store for all databases in the Bird Dev benchmark. ``Baseline'' corresponds to the non-procedural agent (NP-0), and ``PA'' corresponds to the full Procedural Agent (P-3). Error bars indicate $\pm$1 standard deviation. ``Evidence Coverage'' denotes the proportion of test questions that have at least one corresponding question in the memory store whose annotated evidence field has cosine similarity $\geq$ 0.9 to that of the test question.}
    \label{fig:all_learning_curves}
\end{figure}

We present a similar figure to Figure~\ref{fig:learning_curves} for all databases in Figure~\ref{fig:all_learning_curves}.

\clearpage
\section{Sample Trajectories}\label{appdx:trajectories}
\
\subsection{Non-procedural agent with Memory Level 0 -  Offline Sample Trajectory}

% Define colors for different agents
\definecolor{humancolor}{RGB}{70,130,180} % Steel Blue
\definecolor{sqlagentcolor}{RGB}{60,179,113} % Medium Sea Green
\definecolor{toolcallcolor}{RGB}{255,165,0} % Orange
\definecolor{toolresponsecolor}{RGB}{186,85,211} % Medium Orchid
\definecolor{thoughtcolor}{RGB}{169,169,169} % Dark Gray
\definecolor{codebackground}{RGB}{240,240,240} % Light Gray

% Custom environment for messages
\newenvironment{agentmessage}[2]{%
    \begin{tcolorbox}[
        colback=white,
        colframe=#1,
        arc=2mm,
        title={\textbf{#2}},
        fonttitle=\bfseries\color{white},
        coltitle=#1,
        enhanced,
        attach boxed title to top left={yshift=-2mm, xshift=3mm},
        boxed title style={sharp corners, boxrule=0pt}
    ]
}{%
    \end{tcolorbox}
}

% Custom environment for code blocks
\lstdefinestyle{sqlcode}{
    backgroundcolor=\color{codebackground},
    basicstyle=\ttfamily\small,
    breaklines=true,
    captionpos=b,
    keepspaces=true,
    showspaces=false,
    showstringspaces=false,
    showtabs=false,
    tabsize=2,
    frame=single,
    framerule=0.5pt,
    rulecolor=\color{black!30}
}

% Human to SQL agent
\begin{agentmessage}{humancolor}{Human $\rightarrow$ SQL agent}
Generate a SQL query for the following question. I have provided name of the 
database that is relevant to this query.

\textbf{question:}
Among the accounts who have loan validity more than 12 months, list out the accounts that have the 
highest approved amount and have account opening date in 1993.

\textbf{database name:}
financial
\end{agentmessage}

\vspace{-0.1cm}

% SQL agent tool call
\begin{agentmessage}{toolcallcolor}{SQL agent Tool Call: generate\_sql}
\begin{lstlisting}[style=sqlcode]
{
    'question': 'Among the accounts who have loan validity more than 12 months,
list out the accounts that have the highest approved amount and have account 
opening date in 1993.',
    'database_name': 'financial'
}
\end{lstlisting}
Database name: financial\\
../../datasets/bird/dataset/dev/dev/\_databases/financial/database/\_description/account.csv ***\\
../../datasets/bird/dataset/dev/dev/\_databases/financial/database/\_description/card.csv ***\\
../../datasets/bird/dataset/dev/dev/\_databases/financial/database/\_description/client.csv ***\\
../../datasets/bird/dataset/dev/dev/\_databases/financial/database/\_description/disp.csv ***\\
../../datasets/bird/dataset/dev/dev/\_databases/financial/database/\_description/district.csv ***\\
../../datasets/bird/dataset/dev/dev/\_databases/financial/database/\_description/loan.csv ***\\
../../datasets/bird/dataset/dev/dev/\_databases/financial/database/\_description/order.csv ***\\
../../datasets/bird/dataset/dev/dev/\_databases/financial/database/\_description/trans.csv ***\\
Using examples
\end{agentmessage}

\vspace{-0.1cm}

% Similar example section
\begin{agentmessage}{toolresponsecolor}{Similar Example}
\textbf{Question:} Among the accounts who have loan validity more than 24 months, list 
out the accounts that have the lowest approved amount and have account opening date
before 1997.

\textbf{SQL:} 
\begin{lstlisting}[style=sqlcode]
SELECT a.account_id
FROM loan l
JOIN account a ON l.account_id = a.account_id
WHERE l.duration > 24 AND STRFTIME('%Y', a.date) < '1997'
ORDER BY l.amount ASC
LIMIT 1
\end{lstlisting}
\end{agentmessage}

\vspace{-0.1cm}

% SQL agent tool response
\begin{agentmessage}{toolresponsecolor}{SQL agent Tool Response: generate\_sql}
SQL query generated! Generate SQL is 
\begin{lstlisting}[style=sqlcode]
SELECT a.account_id
FROM loan l
JOIN account a ON l.account_id = a.account_id
WHERE l.duration > 12 AND STRFTIME('%Y', a.date) = '1993'
ORDER BY l.amount DESC
LIMIT 1;
\end{lstlisting}
\end{agentmessage}

\vspace{-0.1cm}

% SQL agent to Human
\begin{agentmessage}{sqlagentcolor}{SQL agent $\rightarrow$ Human}
Here is the SQL query to find the accounts with the highest 
approved amount among those with loan validity more than 12 months and account 
opening date in 1993.

\begin{lstlisting}[style=sqlcode]
SELECT a.account_id
FROM loan l
JOIN account a ON l.account_id = a.account_id
WHERE l.duration > 12 AND STRFTIME('%Y', a.date) = '1993'
ORDER BY l.amount DESC
LIMIT 1;
\end{lstlisting}
\end{agentmessage}

\vspace{-0.1cm}

% Query result
\begin{agentmessage}{humancolor}{Query Result}
\begin{lstlisting}[style=sqlcode]
[(10451,)]
\end{lstlisting}
Output match!
\end{agentmessage}

\newpage
\subsection{Non-procedural agent with Memory Level 0 Online Sample Trajectory}

% Define colors for different agents
\definecolor{humancolor}{RGB}{70,130,180} % Steel Blue
\definecolor{sqlagentcolor}{RGB}{60,179,113} % Medium Sea Green
\definecolor{toolcallcolor}{RGB}{255,165,0} % Orange
\definecolor{toolresponsecolor}{RGB}{186,85,211} % Medium Orchid
\definecolor{thoughtcolor}{RGB}{169,169,169} % Dark Gray
\definecolor{codebackground}{RGB}{240,240,240} % Light Gray
\definecolor{humanproxycolor}{RGB}{205,92,92} % Indian Red

% Custom environment for messages
\renewenvironment{agentmessage}[2]{%
    \begin{tcolorbox}[
        colback=white,
        colframe=#1,
        arc=2mm,
        title={\textbf{#2}},
        fonttitle=\bfseries\color{white},
        coltitle=#1,
        enhanced,
        attach boxed title to top left={yshift=-2mm, xshift=3mm},
        boxed title style={sharp corners, boxrule=0pt}
    ]
}{%
    \end{tcolorbox}
}

% Custom environment for code blocks
\lstdefinestyle{sqlcode}{
    backgroundcolor=\color{codebackground},
    basicstyle=\ttfamily\small,
    breaklines=true,
    captionpos=b,
    keepspaces=true,
    showspaces=false,
    showstringspaces=false,
    showtabs=false,
    tabsize=2,
    frame=single,
    framerule=0.5pt,
    rulecolor=\color{black!30}
}

% \begin{center}
%     \LARGE\textbf{SQL Agent Conversation Log}
% \end{center}

% Human to SQL agent
\begin{agentmessage}{humancolor}{Human $\rightarrow$ SQL agent}
Generate a SQL query for the following question. I have provided name of the 
database that is relevant to this query.

\textbf{question:}
In the branch where the second-highest number of crimes were committed in 1995 occurred, how many male 
clients are there?

\textbf{database name:}
financial
\end{agentmessage}

\vspace{-0.1cm}

% SQL agent tool call
\begin{agentmessage}{toolcallcolor}{SQL agent Tool Call: generate\_sql}
\begin{lstlisting}[style=sqlcode]
{
    'question': 'In the branch where the second-highest number of crimes were 
committed in 1995 occurred, how many male clients are there?',
    'database_name': 'financial'
}
\end{lstlisting}
Database name: financial\\
../../datasets/bird/dataset/dev/dev/\_databases/financial/database/\_description/account.csv ***\\
../../datasets/bird/dataset/dev/dev/\_databases/financial/database/\_description/card.csv ***\\
../../datasets/bird/dataset/dev/dev/\_databases/financial/database/\_description/client.csv ***\\
../../datasets/bird/dataset/dev/dev/\_databases/financial/database/\_description/disp.csv ***\\
../../datasets/bird/dataset/dev/dev/\_databases/financial/database/\_description/district.csv ***\\
../../datasets/bird/dataset/dev/dev/\_databases/financial/database/\_description/loan.csv ***\\
../../datasets/bird/dataset/dev/dev/\_databases/financial/database/\_description/order.csv ***\\
../../datasets/bird/dataset/dev/dev/\_databases/financial/database/\_description/trans.csv ***\\
Not using examples
\end{agentmessage}

\vspace{-0.1cm}

% SQL agent tool response
\begin{agentmessage}{toolresponsecolor}{SQL agent Tool Response: generate\_sql}
SQL query generated! Generate SQL is 
\begin{lstlisting}[style=sqlcode]
SELECT COUNT(c.client_id) 
FROM client c
JOIN disp d ON c.client_id = d.client_id
JOIN account a ON d.account_id = a.account_id
JOIN district dis ON a.district_id = dis.district_id
WHERE c.gender = 'male' 
AND dis.district_id = (
    SELECT district_id 
    FROM district 
    ORDER BY A15 DESC 
    LIMIT 1 OFFSET 1
);
\end{lstlisting}
\end{agentmessage}

\vspace{-0.1cm}

% SQL agent to Human proxy agent
\begin{agentmessage}{sqlagentcolor}{SQL agent $\rightarrow$ Human proxy agent}
Could you please review the generated SQL query for correctness?

\begin{lstlisting}[style=sqlcode]
SELECT COUNT(c.client_id) 
FROM client c
JOIN disp d ON c.client_id = d.client_id
JOIN account a ON d.account_id = a.account_id
JOIN district dis ON a.district_id = dis.district_id
WHERE c.gender = 'male' 
AND dis.district_id = (
    SELECT district_id 
    FROM district 
    ORDER BY A15 DESC 
    LIMIT 1 OFFSET 1
);
\end{lstlisting}
\end{agentmessage}

\vspace{-0.1cm}

% Human proxy agent tool call
\begin{agentmessage}{toolcallcolor}{Human proxy agent Tool Call: evaluate\_sql}
\begin{lstlisting}[style=sqlcode]
{
    'generated_sql': "SELECT COUNT(c.client_id) \nFROM client c\nJOIN disp d ON
c.client_id = d.client_id\nJOIN account a ON d.account_id = a.account_id\nJOIN 
district dis ON a.district_id = dis.district_id\nWHERE c.gender = 'male' \nAND 
dis.district_id = (\n    SELECT district_id \n    FROM district \n    ORDER BY 
A15 DESC \n    LIMIT 1 OFFSET 1\n);"
}
\end{lstlisting}
../../datasets/bird/dataset/dev/dev\_databases\textbackslash financial\textbackslash financial.sqlite\\
Output match fail!
\end{agentmessage}

\vspace{-0.1cm}

% Human proxy agent tool response
\begin{agentmessage}{toolresponsecolor}{Human proxy agent Tool Response: evaluate\_sql}
\begin{lstlisting}[style=sqlcode]
{
    'test_sql': "SELECT COUNT(c.client_id) \nFROM client c\nJOIN disp d ON 
c.client_id = d.client_id\nJOIN account a ON d.account_id = a.account_id\nJOIN 
district dis ON a.district_id = dis.district_id\nWHERE c.gender = 'male' \nAND 
dis.district_id = (\n    SELECT district_id \n    FROM district \n    ORDER BY 
A15 DESC \n    LIMIT 1 OFFSET 1\n);",
    'gold_sql': "SELECT COUNT(T1.client_id) FROM client AS T1 INNER JOIN 
district AS T2 ON T1.district_id = T2.district_id WHERE T1.gender = 'M' AND 
T2.A15 = (SELECT T3.A15 FROM district AS T3 ORDER BY T3.A15 DESC LIMIT 1, 1)",
    'outcome': 'output-match-fail'
}
\end{lstlisting}
\end{agentmessage}

\vspace{-0.1cm}

% Human proxy agent thought
\begin{agentmessage}{thoughtcolor}{Human proxy agent $\rightarrow$ Human proxy agent (Thought)}
The generated SQL query does not match the expected output. I need
to analyze the differences and provide feedback to the SQL agent on how to 
correct the query.
\end{agentmessage}

\vspace{-0.1cm}

% Human proxy agent thought
\begin{agentmessage}{thoughtcolor}{Human proxy agent $\rightarrow$ Human proxy agent (Thought)}
The generated SQL query does not produce the correct output. The 
main issues are:

1. \textbf{Gender Filtering}: The query uses \texttt{c.gender = 
'male'}, but the correct filter should be \texttt{c.gender = 'M'} to match the 
expected format in the database.

2. \textbf{District Identification}: The 
subquery correctly identifies the district with the second-highest number of 
crimes using \texttt{LIMIT 1 OFFSET 1}, which is equivalent to \texttt{LIMIT 1, 1}. However, 
the subquery should select \texttt{A15} instead of \texttt{district\_id} to match the number 
of crimes.

3. \textbf{Table Joins}: The query unnecessarily joins the \texttt{disp} and 
\texttt{account} tables, which are not needed for this task. The \texttt{client} table can be
directly joined with the \texttt{district} table using \texttt{district\_id}.

The SQL agent
should adjust the query to address these issues.
\end{agentmessage}

\vspace{-0.1cm}

% Human proxy agent to SQL agent
\begin{agentmessage}{humanproxycolor}{Human proxy agent $\rightarrow$ SQL agent}
The generated SQL query has a few issues that need to be addressed:

1. \textbf{Gender Filtering}: The query uses \texttt{c.gender = 'male'}, but it should use 
\texttt{c.gender = 'M'} to match the expected format in the database.

2. \textbf{District
Identification}: The subquery should select \texttt{A15} instead of \texttt{district\_id} to 
correctly identify the district with the second-highest number of crimes.

3. \textbf{Table Joins}: The joins with the \texttt{disp} and \texttt{account} tables are unnecessary
for this task. You can directly join the \texttt{client} table with the \texttt{district} 
table using \texttt{district\_id}.

Please adjust the query accordingly to address 
these issues.
\end{agentmessage}

\vspace{-0.1cm}

% SQL agent tool call
\begin{agentmessage}{toolcallcolor}{SQL agent Tool Call: refine\_sql}
\begin{lstlisting}[style=sqlcode]
{
    'question': 'In the branch where the second-highest number of crimes were 
committed in 1995 occurred, how many male clients are there?',
    'generated_sql': "SELECT COUNT(c.client_id) \nFROM client c\nJOIN disp d ON
c.client_id = d.client_id\nJOIN account a ON d.account_id = a.account_id\nJOIN 
district dis ON a.district_id = dis.district_id\nWHERE c.gender = 'male' \nAND 
dis.district_id = (\n    SELECT district_id \n    FROM district \n    ORDER BY 
A15 DESC \n    LIMIT 1 OFFSET 1\n);",
    'feedback': "The generated SQL query has a few issues that need to be 
addressed:\n\n1. **Gender Filtering**: The query uses `c.gender = 'male'`, but 
it should use `c.gender = 'M'` to match the expected format in the 
database.\n\n2. **District Identification**: The subquery should select `A15` 
instead of `district_id` to correctly identify the district with the 
second-highest number of crimes.\n\n3. **Table Joins**: The joins with the 
`disp` and `account` tables are unnecessary for this task. You can directly 
join the `client` table with the `district` table using `district_id`.",
    'database_name': 'financial'
}
\end{lstlisting}
\end{agentmessage}

\vspace{-0.1cm}

% SQL agent tool response
\begin{agentmessage}{toolresponsecolor}{SQL agent Tool Response: refine\_sql}
SQL refined! Refined SQL is 
\begin{lstlisting}[style=sqlcode]
SELECT COUNT(c.client_id) 
FROM client c
JOIN district dis ON c.district_id = dis.district_id
WHERE c.gender = 'M' 
AND dis.district_id = (
    SELECT district_id 
    FROM district 
    ORDER BY A15 DESC 
    LIMIT 1 OFFSET 1
);
\end{lstlisting}
\end{agentmessage}

\vspace{-0.1cm}

% SQL agent to Human proxy agent
\begin{agentmessage}{sqlagentcolor}{SQL agent $\rightarrow$ Human proxy agent}
Could you please review the refined SQL query for correctness?

\begin{lstlisting}[style=sqlcode]
SELECT COUNT(c.client_id) 
FROM client c
JOIN district dis ON c.district_id = dis.district_id
WHERE c.gender = 'M' 
AND dis.district_id = (
    SELECT district_id 
    FROM district 
    ORDER BY A15 DESC 
    LIMIT 1 OFFSET 1
);
\end{lstlisting}
\end{agentmessage}

\vspace{-0.1cm}

% Human proxy agent tool call
\begin{agentmessage}{toolcallcolor}{Human proxy agent Tool Call: evaluate\_sql}
\begin{lstlisting}[style=sqlcode]
{
    'generated_sql': "SELECT COUNT(c.client_id) \nFROM client c\nJOIN district 
dis ON c.district_id = dis.district_id\nWHERE c.gender = 'M' \nAND 
dis.district_id = (\n    SELECT district_id \n    FROM district \n    ORDER BY 
A15 DESC \n    LIMIT 1 OFFSET 1\n);\n"
}
\end{lstlisting}
../../datasets/bird/dataset/dev/dev\_databases\textbackslash financial\textbackslash financial.sqlite\\
Output match!
\end{agentmessage}

\vspace{-0.1cm}

% Human proxy agent tool response
\begin{agentmessage}{toolresponsecolor}{Human proxy agent Tool Response: evaluate\_sql}
\begin{lstlisting}[style=sqlcode]
{
    'test_sql': "SELECT COUNT(c.client_id) \nFROM client c\nJOIN district dis 
ON c.district_id = dis.district_id\nWHERE c.gender = 'M' \nAND dis.district_id 
= (\n    SELECT district_id \n    FROM district \n    ORDER BY A15 DESC \n    
LIMIT 1 OFFSET 1\n);\n",
    'gold_sql': "SELECT COUNT(T1.client_id) FROM client AS T1 INNER JOIN 
district AS T2 ON T1.district_id = T2.district_id WHERE T1.gender = 'M' AND 
T2.A15 = (SELECT T3.A15 FROM district AS T3 ORDER BY T3.A15 DESC LIMIT 1, 1)",
    'outcome': 'output-match'
}
\end{lstlisting}
\end{agentmessage}

\vspace{-0.1cm}

% Human proxy agent to SQL agent

\begin{agentmessage}{humanproxycolor}{Human proxy agent $\rightarrow$ SQL agent}
The refined SQL query is now correct and produces the expected output. Great 
job on making the necessary adjustments!
\end{agentmessage}

\vspace{-0.1cm}

% SQL agent to Human
\begin{agentmessage}{sqlagentcolor}{SQL agent $\rightarrow$ Human}
The SQL query has been successfully generated and verified. Here 
is the final query for your request:

\begin{lstlisting}[style=sqlcode]
SELECT COUNT(c.client_id) 
FROM client c
JOIN district dis ON c.district_id = dis.district_id
WHERE c.gender = 'M' 
AND dis.district_id = (
    SELECT district_id 
    FROM district 
    ORDER BY A15 DESC 
    LIMIT 1 OFFSET 1
);
\end{lstlisting}
\end{agentmessage}

\vspace{-0.1cm}

% Query result
\begin{agentmessage}{humancolor}{Query Result}
\begin{lstlisting}[style=sqlcode]
[(96,)]
\end{lstlisting}
Output match!
\end{agentmessage}

\vspace{-0.1cm}

% Knowledge summary
\begin{agentmessage}{thoughtcolor}{Knowledge Summary}
In the context of determining the number of male clients in the branch where the second-highest number of crimes were committed in 1995, the following knowledge gaps were identified and addressed:

1. \textbf{Understanding of Crime Data}: The number of crimes committed in 1995 is stored in the \texttt{A15} column of the \texttt{district} table. This column should be used to determine the branch with the second-highest number of crimes.

2. \textbf{Branch Identification}: The \texttt{district\_id} column is used to identify branches across different tables. This column is crucial for linking the \texttt{district} table with the \texttt{client} table to find the correct branch.

3. \textbf{Gender Filtering}: The \texttt{gender} column in the \texttt{client} table is used to filter for male clients. The value to filter for male clients is \texttt{'M'}.

4. \textbf{Joining Tables}: To find the number of male clients in a specific branch, it is necessary to join the \texttt{client} table with the \texttt{district} table using the \texttt{district\_id} column.

5. \textbf{Handling of Rankings}: To find the second-highest number of crimes, it is necessary to order the districts by the number of crimes in descending order and then select the second entry.

By addressing these gaps, the correct SQL query can be constructed to accurately answer the user's question.
\end{agentmessage}

\newpage
\subsection{Procedural Agent with Memory Level 3 - Offline Sample Trajectory}
\vspace{2mm}
% \geometry{margin=1in}

% Define colors for different agents
\definecolor{humancolor}{RGB}{70,130,180} % Steel Blue
\definecolor{sqlagentcolor}{RGB}{60,179,113} % Medium Sea Green
\definecolor{toolcallcolor}{RGB}{255,165,0} % Orange
\definecolor{toolresponsecolor}{RGB}{186,85,211} % Medium Orchid
\definecolor{thoughtcolor}{RGB}{169,169,169} % Dark Gray
\definecolor{codebackground}{RGB}{240,240,240} % Light Gray

% Custom environment for messages
\renewenvironment{agentmessage}[2]{%
    \begin{tcolorbox}[
        breakable,
        colback=white,
        colframe=#1,
        arc=2mm,
        title={\textbf{#2}},
        fonttitle=\bfseries\color{white},
        coltitle=#1,
        enhanced,
        attach boxed title to top left={yshift=-2mm, xshift=3mm},
        boxed title style={sharp corners, boxrule=0pt}
    ]
}{%
    \end{tcolorbox}
}

% Custom environment for code blocks
\lstdefinestyle{sqlcode}{
    backgroundcolor=\color{codebackground},
    basicstyle=\ttfamily\small,
    breaklines=true,
    captionpos=b,
    keepspaces=true,
    showspaces=false,
    showstringspaces=false,
    showtabs=false,
    tabsize=2,
    frame=single,
    framerule=0.5pt,
    rulecolor=\color{black!30}
}

% \begin{document}

% \begin{center}
%     \LARGE\textbf{Agent Conversation Log: Procedural Agent at Test Time (Post-learning)}
% \end{center}
\vspace{-1em}
% Human to SQL agent
\begin{agentmessage}{humancolor}{Human $\rightarrow$ SQL agent}
Generate a SQL query for the following question. I have provided name of the 
database that is relevant to this query.

\textbf{question:}
Among the accounts who have loan validity more than 12 months, list out the accounts that have the 
highest approved amount and have account opening date in 1993.

\textbf{database name:}
financial
\end{agentmessage}

% \vspace{0.5cm}
\vspace{-0.1cm}

% SQL agent tool call
\begin{agentmessage}{toolcallcolor}{SQL agent Tool Call: find\_memory}
\begin{lstlisting}[style=sqlcode]
{
    'query_string': 'accounts with loan validity more than 12 months and 
highest approved amount',
    'memory_type': 'similar_question'
}
\end{lstlisting}
\end{agentmessage}

\vspace{-0.1cm}

% Tool response
\begin{agentmessage}{toolresponsecolor}{SQL agent Tool Response: find\_memory}
\small
Index: Among the accounts who have loan validity more than 24 months, list out
the accounts that have the lowest approved amount and have account opening date
before 1997. \\
\textbf{SQL:} 
\begin{lstlisting}[style=sqlcode]
SELECT a.account_id
FROM loan l
JOIN account a ON l.account_id = a.account_id
WHERE l.duration > 24 AND STRFTIME('%Y', a.date) < '1997'
ORDER BY l.amount ASC
LIMIT 1
\end{lstlisting}

\textbf{Knowledge:} 
1. \textbf{Ordering and Limiting Results}: When you need to find the minimum or maximum value in a 
dataset, use ORDER BY to sort the results and LIMIT to restrict the output to 
the desired number of records. This is a standard SQL practice for efficiently 
retrieving extreme values.

2. \textbf{Date Filtering with STRFTIME}: Use the STRFTIME function to extract specific components 
(like year, month, day) from date columns for accurate filtering. This is particularly useful when you need 
to filter records based on a specific time period.

3. \textbf{Schema Understanding}: Familiarize yourself with the columns in each table and their 
data types. This knowledge is crucial for constructing accurate SQL queries 
that meet specific requirements, such as filtering based on date or retrieving 
specific columns.

[Additional memory entries omitted for brevity]
\end{agentmessage}

% \vspace{0.5cm}

% SQL agent thought
\begin{agentmessage}{thoughtcolor}{SQL agent $\rightarrow$ SQL agent (Thought)}
The similar question found involves filtering accounts based on 
loan duration and account opening date, but it focuses on the lowest approved 
amount and a different year. However, it provides useful insights on how to 
structure the query:

1. \textbf{Join the \texttt{loan} and \texttt{account} tables}: This is 
necessary to access both loan details and account opening dates.
2. \textbf{Filter by loan duration}: Use a condition to filter loans with a duration greater 
than 12 months.
3. \textbf{Filter by account opening date}: Use the \texttt{STRFTIME} 
function to extract the year from the account opening date and filter for the 
year 1993.
4. \textbf{Order by approved amount}: Use \texttt{ORDER BY} to sort the results
by the approved amount in descending order to find the highest.
5. \textbf{Limit the results}: Use \texttt{LIMIT} to ensure only the account with the highest approved 
amount is returned.

Next, I will search for any relevant subtasks or 
database facts that could assist in constructing the SQL query.
\end{agentmessage}

% \vspace{0.5cm}

% Additional tool calls and responses would go here
% [Additional tool calls and responses omitted for brevity]

% \vspace{0.5cm}

% Final SQL agent thought
\begin{agentmessage}{thoughtcolor}{SQL agent $\rightarrow$ SQL agent (Final Thought)}
The SQL query aligns with the information from the memories:
\begin{itemize}
\item The join between \texttt{loan} and \texttt{account} tables is correctly implemented.
\item The filter for loan duration greater than 12 months is correctly applied.
\item The use of \texttt{STRFTIME} to filter accounts opened in 1993 is consistent with the memory.
\item The ordering by approved amount in descending order and limiting the results to 1 is correctly implemented.
\end{itemize}

I am confident that this SQL query is accurate and ready to be shared with the human.
\end{agentmessage}

% \vspace{0.5cm}

% SQL agent response to human
\begin{agentmessage}{sqlagentcolor}{SQL agent $\rightarrow$ Human}
Here is the SQL query to find the accounts with loan validity more
than 12 months, the highest approved amount, and account opening date in 
1993.

\begin{lstlisting}[style=sqlcode]
SELECT a.account_id
FROM loan l
JOIN account a ON l.account_id = a.account_id
WHERE l.duration > 12 AND STRFTIME('%Y', a.date) = '1993'
ORDER BY l.amount DESC
LIMIT 1;
\end{lstlisting}
\end{agentmessage}

% \vspace{0.5cm}

% Query result
\begin{agentmessage}{humancolor}{Query Result}
\begin{lstlisting}[style=sqlcode]
[(10451,)]
\end{lstlisting}
Output match!
\end{agentmessage}

% \end{document}

\newpage
\subsection{Procedural Agent with Memory Level 3 - Online Sample Trajectory}

% \documentclass{article}
% \usepackage[utf8]{inputenc}
% \usepackage{xcolor}
% \usepackage{tcolorbox}
% \usepackage{listings}
% \usepackage{fontawesome}
% \usepackage{geometry}
% \geometry{margin=1in}

% Define colors for different agents
\definecolor{humancolor}{RGB}{70,130,180} % Steel Blue
\definecolor{sqlagentcolor}{RGB}{60,179,113} % Medium Sea Green
\definecolor{toolcallcolor}{RGB}{255,165,0} % Orange
\definecolor{toolresponsecolor}{RGB}{186,85,211} % Medium Orchid
\definecolor{thoughtcolor}{RGB}{169,169,169} % Dark Gray
\definecolor{codebackground}{RGB}{240,240,240} % Light Gray

% Custom environment for messages
\renewenvironment{agentmessage}[2]{%
    \begin{tcolorbox}[
        colback=white,
        colframe=#1,
        arc=2mm,
        title={\textbf{#2}},
        fonttitle=\bfseries\color{white},
        coltitle=#1,
        enhanced,
        attach boxed title to top left={yshift=-2mm, xshift=3mm},
        boxed title style={sharp corners, boxrule=0pt}
    ]
}{%
    \end{tcolorbox}
}

% Custom environment for code blocks
\lstdefinestyle{sqlcode}{
    backgroundcolor=\color{codebackground},
    basicstyle=\ttfamily\small,
    breaklines=true,
    captionpos=b,
    keepspaces=true,
    showspaces=false,
    showstringspaces=false,
    showtabs=false,
    tabsize=2,
    frame=single,
    framerule=0.5pt,
    rulecolor=\color{black!30}
}

% \begin{document}

% \begin{center}
%         \LARGE\textbf{Agent Conversation Log: Procedural Agent During Online Learning}
% \end{center}

% Human to SQL agent
\begin{agentmessage}{humancolor}{Human $\rightarrow$ SQL agent}
Generate a SQL query for the following question. I have provided name of the 
database that is relevant to this query.

\textbf{question:}
In the branch where the second-highest number of crimes were committed in 1995 occurred, how many male 
clients are there?

\textbf{database name:}
financial
\end{agentmessage}

%\vspace{-0.1cm}

% SQL agent thought
\begin{agentmessage}{thoughtcolor}{SQL agent $\rightarrow$ SQL agent (Thought)}
To solve this query, I need to follow these steps:

1. Identify the branch (district) with the second-highest number of crimes committed in 
1995. This information is available in the 'district' table, specifically in 
the 'A15' column.

2. Once the district is identified, I need to find the 
number of male clients in that district. This requires joining the 'client' 
table with the 'district' table using the 'district\_id'.

3. Filter the 
clients by gender to count only male clients.

4. Assemble the SQL query 
using these steps and ensure it adheres to the database schema.

Next, I will
start by identifying the district with the second-highest number of crimes in 
1995.
\end{agentmessage}

%\vspace{-0.1cm}

% SQL agent thought
\begin{agentmessage}{thoughtcolor}{SQL agent $\rightarrow$ SQL agent (Thought)}
To find the district with the second-highest number of crimes in 
1995, I will use the 'district' table and order the districts by the 'A15' 
column in descending order. I will then select the second entry from this 
ordered list. 

The SQL query for this step would look like 
this:

\begin{lstlisting}[style=sqlcode]
SELECT district_id FROM district ORDER BY A15 DESC LIMIT 1 OFFSET 1;
\end{lstlisting}

This query will give me the district\_id of the branch with 
the second-highest number of crimes in 1995. 

Next, I will plan how to count
the number of male clients in this district.
\end{agentmessage}

%\vspace{-0.1cm}

% SQL agent thought
\begin{agentmessage}{thoughtcolor}{SQL agent $\rightarrow$ SQL agent (Thought)}
Now that I have the district\_id with the second-highest number of 
crimes in 1995, I need to count the number of male clients in that district. 

To do this, I will join the 'client' table with the 'district' table using 
the 'district\_id'. I will then filter the clients by gender to count only male 
clients. 

The SQL query for this step would look like 
this:

\begin{lstlisting}[style=sqlcode]
SELECT COUNT(*) FROM client WHERE gender = 'male' AND 
district_id = (SELECT district_id FROM district ORDER BY A15 DESC LIMIT 1 
OFFSET 1);
\end{lstlisting}

This query will give me the count of male clients in the 
district with the second-highest number of crimes in 1995. 

I will now send 
this query to the database expert agent for feedback.
\end{agentmessage}

%\vspace{-0.1cm}

% SQL agent to Human proxy agent
\begin{agentmessage}{sqlagentcolor}{SQL agent $\rightarrow$ Human proxy agent}
Please review the following SQL query to ensure it correctly 
counts the number of male clients in the district with the second-highest 
number of crimes in 1995.

\begin{lstlisting}[style=sqlcode]
SELECT COUNT(*) FROM client WHERE gender = 'male' AND 
district_id = (SELECT district_id FROM district ORDER BY A15 DESC LIMIT 1 
OFFSET 1);
\end{lstlisting}
\end{agentmessage}

%\vspace{-0.1cm}

% Human proxy agent tool call
\begin{agentmessage}{toolcallcolor}{Human proxy agent Tool Call: evaluate\_sql}
\begin{lstlisting}[style=sqlcode]
{
    'generated_sql': "SELECT COUNT(*) FROM client WHERE gender = 'male' AND 
district_id = (SELECT district_id FROM district ORDER BY A15 DESC LIMIT 1 
OFFSET 1);"
}
\end{lstlisting}
../../datasets/bird/dataset/dev/dev\_databases\textbackslash financial\textbackslash financial.sqlite\\
Output match fail!
\end{agentmessage}

%\vspace{-0.1cm}

% Human proxy agent tool response
\begin{agentmessage}{toolresponsecolor}{Human proxy agent Tool Response: evaluate\_sql}
\begin{lstlisting}[style=sqlcode]
{
    'test_sql': "SELECT COUNT(*) FROM client WHERE gender = 'male' AND 
district_id = (SELECT district_id FROM district ORDER BY A15 DESC LIMIT 1 
OFFSET 1);",
    'gold_sql': "SELECT COUNT(T1.client_id) FROM client AS T1 INNER JOIN 
district AS T2 ON T1.district_id = T2.district_id WHERE T1.gender = 'M' AND 
T2.A15 = (SELECT T3.A15 FROM district AS T3 ORDER BY T3.A15 DESC LIMIT 1, 1)",
    'outcome': 'output-match-fail'
}
\end{lstlisting}
\end{agentmessage}

%\vspace{-0.1cm}

% Human proxy agent thought
\begin{agentmessage}{thoughtcolor}{Human proxy agent $\rightarrow$ Human proxy agent (Thought)}
The SQL query provided by the SQL agent does not match the 
expected output. The main issues are:

1. \textbf{Gender Filtering}: The query 
uses \texttt{gender = 'male'}, but the correct filter should be \texttt{gender = 'M'} as per 
the database schema.

2. \textbf{Subquery for District}: The subquery correctly 
identifies the district with the second-highest number of crimes, but it should
return the \texttt{A15} value instead of \texttt{district\_id} to match the expert 
solution.

3. \textbf{Table Join}: The query does not join the \texttt{client} and 
\texttt{district} tables, which is necessary to ensure the correct district is being 
referenced.

I will provide feedback to the SQL agent to address these 
issues.
\end{agentmessage}

%\vspace{-0.1cm}

% Human proxy agent to SQL agent
\begin{agentmessage}{humanproxycolor}{Human proxy agent $\rightarrow$ SQL agent}
The SQL query needs some adjustments to correctly count the number of male 
clients in the district with the second-highest number of crimes in 1995:

1. \textbf{Gender Filtering}: Use \texttt{gender = 'M'} instead of \texttt{gender = 'male'} to match 
the database schema.

2. \textbf{Subquery for District}: Ensure the subquery 
returns the \texttt{A15} value for the number of crimes, not the \texttt{district\_id}. This 
is crucial for identifying the correct district.

3. \textbf{Table Join}: 
Implement an \texttt{INNER JOIN} between the \texttt{client} and \texttt{district} tables on 
\texttt{district\_id} to ensure the correct district is being referenced.

Please 
revise the query with these points in mind.
\end{agentmessage}

%\vspace{-0.1cm}

% SQL agent thought
\begin{agentmessage}{thoughtcolor}{SQL agent $\rightarrow$ SQL agent (Thought)}
To revise the SQL query, I need to make the following changes 
based on the feedback:

1. Change the gender filter to \texttt{gender = 'M'} to 
match the database schema.

2. Correct the subquery to ensure it returns the 
\texttt{district\_id} of the district with the second-highest number of crimes in 
1995.

3. Implement an \texttt{INNER JOIN} between the \texttt{client} and \texttt{district} 
tables on \texttt{district\_id} to ensure the correct district is being 
referenced.

The revised SQL query should look like this:

\begin{lstlisting}[style=sqlcode]
SELECT COUNT(*) 
FROM client 
INNER JOIN district ON client.district_id = district.district_id 
WHERE client.gender = 'M' 
AND district.district_id = (SELECT district_id FROM district ORDER BY A15 DESC LIMIT 1 OFFSET 1);
\end{lstlisting}

I will now send this revised query to the database expert agent for
feedback.
\end{agentmessage}

%\vspace{-0.1cm}

% SQL agent to Human proxy agent
\begin{agentmessage}{sqlagentcolor}{SQL agent $\rightarrow$ Human proxy agent}
Please review the revised SQL query to ensure it correctly counts 
the number of male clients in the district with the second-highest number of 
crimes in 1995.

\begin{lstlisting}[style=sqlcode]
SELECT COUNT(*) 
FROM client 
INNER JOIN district ON client.district_id = district.district_id 
WHERE client.gender = 'M' 
AND district.district_id = (SELECT district_id FROM district ORDER BY A15 DESC LIMIT 1 OFFSET 1);
\end{lstlisting}
\end{agentmessage}

%\vspace{-0.1cm}

% Human proxy agent tool call
\begin{agentmessage}{toolcallcolor}{Human proxy agent Tool Call: evaluate\_sql}
\begin{lstlisting}[style=sqlcode]
{
    'generated_sql': "SELECT COUNT(*) \nFROM client \nINNER JOIN district ON 
client.district_id = district.district_id \nWHERE client.gender = 'M' \nAND 
district.district_id = (SELECT district_id FROM district ORDER BY A15 DESC 
LIMIT 1 OFFSET 1);"
}
\end{lstlisting}
../../datasets/bird/dataset/dev/dev\_databases\textbackslash financial\textbackslash financial.sqlite\\
Output match!
\end{agentmessage}

%\vspace{-0.1cm}

% Human proxy agent tool response
\begin{agentmessage}{toolresponsecolor}{Human proxy agent Tool Response: evaluate\_sql}
\begin{lstlisting}[style=sqlcode]
{
    'test_sql': "SELECT COUNT(*) \nFROM client \nINNER JOIN district ON 
client.district_id = district.district_id \nWHERE client.gender = 'M' \nAND 
district.district_id = (SELECT district_id FROM district ORDER BY A15 DESC 
LIMIT 1 OFFSET 1);",
    'gold_sql': "SELECT COUNT(T1.client_id) FROM client AS T1 INNER JOIN 
district AS T2 ON T1.district_id = T2.district_id WHERE T1.gender = 'M' AND 
T2.A15 = (SELECT T3.A15 FROM district AS T3 ORDER BY T3.A15 DESC LIMIT 1, 1)",
    'outcome': 'output-match'
}
\end{lstlisting}
\end{agentmessage}

%\vspace{-0.1cm}

% Human proxy agent to SQL agent
\begin{agentmessage}{humanproxycolor}{Human proxy agent $\rightarrow$ SQL agent}
The revised SQL query is now correct and accurately counts the number of male 
clients in the district with the second-highest number of crimes in 1995. Great
job!
\end{agentmessage}

%\vspace{-0.1cm}

% SQL agent to Human
\begin{agentmessage}{sqlagentcolor}{SQL agent $\rightarrow$ Human}
The SQL query has been successfully generated and verified. It 
counts the number of male clients in the district with the second-highest 
number of crimes in 1995. Here is the query:

\begin{lstlisting}[style=sqlcode]
SELECT COUNT(*) 
FROM client 
INNER JOIN district ON client.district_id = district.district_id 
WHERE client.gender = 'M' 
AND district.district_id = (SELECT district_id FROM district ORDER BY A15 DESC LIMIT 1 OFFSET 1);
\end{lstlisting}

This query works by first identifying the district with the second-highest number of crimes in 1995
using a subquery. It then counts the number of male clients in that district by
joining the \texttt{client} and \texttt{district} tables on \texttt{district\_id} and filtering for 
male clients.
\end{agentmessage}

%\vspace{-0.1cm}

% Query result
\begin{agentmessage}{humancolor}{Query Result}
\begin{lstlisting}[style=sqlcode]
[(96,)]
\end{lstlisting}
Output match!

\end{agentmessage}

%\vspace{-0.1cm}

% Human to SQL agent
\begin{agentmessage}{humancolor}{Human $\rightarrow$ SQL agent}
Please summarize what you learned from interacting with the Expert agent. 
Think about it out loud in many steps, and then give me a concise and specific 
summary. 
                        Do these steps sequentially. The output of 
the previous step is used for the subsequent steps.
                        
1. First send yourself a message which describes the mistakes you made. This 
should explicitly state what component of the SQL query was incorrect. You 
should also pinpoint where this error occured in your reasoning.
             2. Second, send yourself a message which details what database knowledge was 
revealed through feedback to correct this mistake. Link this is to the schema. 
Be very specific, and SQL heavy. This may contain information about the 
reasoning you had about the database.
                        3. Third, look 
at the thought generated by step 2. Rephrase the findings as facts. Provide 
some context as to when these facts would be useful in the future.
           
                        The only response that I want back is the set of 
facts. Leave the "generated\_sql" field empty in your response.
\end{agentmessage}

%\vspace{-0.1cm}

% SQL agent thought
\begin{agentmessage}{thoughtcolor}{SQL agent $\rightarrow$ SQL agent (Thought)}
The mistakes I made in the initial SQL query were:

1. \textbf{Gender Filtering}: I used \texttt{gender = 'male'} instead of \texttt{gender = 'M'}. This mistake 
occurred because I assumed the gender values were stored as full words rather 
than single characters.

2. \textbf{Subquery for District}: I incorrectly assumed 
that the subquery should return the \texttt{district\_id} directly without considering 
the need to order by the number of crimes (\texttt{A15}). This error was due to a lack
of attention to the specific requirement of identifying the district with the 
second-highest number of crimes.

3. \textbf{Table Join}: I initially did not 
implement an \texttt{INNER JOIN} between the \texttt{client} and \texttt{district} tables, which is 
necessary to ensure the correct district is being referenced.
\end{agentmessage}

%\vspace{-0.1cm}

% SQL agent thought
\begin{agentmessage}{thoughtcolor}{SQL agent $\rightarrow$ SQL agent (Thought)}
The feedback revealed the following database knowledge:

1. \textbf{Gender Filtering}: The \texttt{client} table stores gender as a single character 
('M' for male, 'F' for female). This is evident from the schema where the 
\texttt{gender} column is of type \texttt{text}.

2. \textbf{Subquery for District}: To find the
district with the second-highest number of crimes in 1995, the subquery must 
order the \texttt{district} table by the \texttt{A15} column in descending order and then 
select the second entry. This ensures the correct district is identified based 
on the number of crimes.

3. \textbf{Table Join}: An \texttt{INNER JOIN} between the 
\texttt{client} and \texttt{district} tables on \texttt{district\_id} is necessary to reference the 
correct district when counting clients. This join operation is crucial for 
combining data from both tables based on the shared \texttt{district\_id}.
\end{agentmessage}

%\vspace{-0.1cm}

% SQL agent thought
\begin{agentmessage}{thoughtcolor}{SQL agent $\rightarrow$ SQL agent (Thought)}
Facts derived from the feedback:

1. \textbf{Gender Representation}: 
In the \texttt{client} table, gender is represented by single characters ('M' for 
male, 'F' for female). This is crucial when filtering clients by gender in SQL 
queries.

2. \textbf{Identifying Districts by Crime Rate}: To find a district 
based on crime rates, use the \texttt{A15} column in the \texttt{district} table. Order the 
districts by this column in descending order to identify districts with the 
highest number of crimes.

3. \textbf{Joining Tables on District ID}: When needing
to reference district-specific data for clients, use an \texttt{INNER JOIN} between 
the \texttt{client} and \texttt{district} tables on \texttt{district\_id}. This join is essential for
combining client data with district data accurately.
\end{agentmessage}

%\vspace{-0.1cm}

% SQL agent to Human
\begin{agentmessage}{sqlagentcolor}{SQL agent $\rightarrow$ Human}
1. \textbf{Gender Representation}: In the \texttt{client} table, gender is 
represented by single characters ('M' for male, 'F' for female). This is 
crucial when filtering clients by gender in SQL queries.

2. \textbf{Identifying 
Districts by Crime Rate}: To find a district based on crime rates, use the 
\texttt{A15} column in the \texttt{district} table. Order the districts by this column in 
descending order to identify districts with the highest number of crimes.

3. \textbf{Joining Tables on District ID}: When needing to reference district-specific 
data for clients, use an \texttt{INNER JOIN} between the \texttt{client} and \texttt{district} 
tables on \texttt{district\_id}. This join is essential for combining client data with 
district data accurately.
\end{agentmessage}

%\vspace{-0.1cm}

% Human to SQL agent
\begin{agentmessage}{humancolor}{Human $\rightarrow$ SQL agent}
Now save the distilled knowledge, one chunk at a time, for future use by using
your save\_memory tool. You can decompose the SQL query and save the chunks as 
sub tasks as useful reusable components in the future. You can also save 
database\_facts, which contain useful information that may be necessary in the 
future in a more general set of tasks. 
                    You can save up to
5 memories. Plan and decide which are most important before you start. Here are
some helpful tips:
                    - Be sure to make the index of the 
memory useful. A useful index is one that will be easy to search for in the 
future. For example, dont just say "join table1 and table2". Say something like
"join table1 and table2 by attribute", where attribute is the common column.
 - When saving a similar subtask, the index should generally describe the 
subtask. Always make the passage contain a chunk of SQL. The index must contain
information about the values filtered for in the SQL if you keep the values in 
the passage. Otherwise, you should replace the filters with placeholders, and 
the index should reflect the generality.
                    - Don't save 
basic SQL knowledge like "how to limit". This is information you already know. 
You could instead say "how to sort for \_\_column name\_\_" and then the passage 
can show that you order and use a desc and limit clause in a SQL query.
      
                    Some examples of what not to do:
                    
'query\_string': 'use of JOIN operations to connect tables',
              'knowledge\_string': 'JOIN operations are used to connect tables based on 
common columns to retrieve related data.',
                    
'memory\_type': 'database\_fact'

                    Reason: This does not
provide any useful information about the database and is just common SQL 
knowledge.

                    'query\_string': 'filter by exact text 
values in SQL',
                    'knowledge\_string': "When constructing 
SQL queries involving the 'frequency' column, it is essential to use the 
exact text values defined in the schema for accurate filtering. This ensures 
the query retrieves the intended results based on the issuance frequency.",
  'memory\_type': 'database\_fact'

                    Reason: You should 
instead save "filter by issuance frequecy" as the index, and the the passage 
should be showing the portion of the SQL query of how to do this. If available,
you should specify what the filter value corresponds to in the index.

      'query\_string': "filter accounts by frequency 'POPLATEK TYDNE'",
         'knowledge\_string': "SELECT * FROM account WHERE frequency = 'POPLATEK 
TYDNE';",
                    'memory\_type': 'similar\_subtask'
         
                    Reason: The query string is not helpful since you would 
need to know to search for POPLATEK TYDNE in the first place. The tricky part 
is knowing what this value semanticly means. Instead the query string should be
"filter accounts for weekly issuance"
\end{agentmessage}

%\vspace{-0.1cm}

% SQL agent tool call
\begin{agentmessage}{toolcallcolor}{SQL agent Tool Call: save\_memory}
\begin{lstlisting}[style=sqlcode]
{
    'query_string': 'filter clients by gender',
    'knowledge_string': "SELECT * FROM client WHERE gender = 'M';",
    'memory_type': 'similar_subtask'
}
\end{lstlisting}
\end{agentmessage}

%\vspace{-0.1cm}

% SQL agent tool response
\begin{agentmessage}{toolresponsecolor}{SQL agent Tool Response: save\_memory}
Saved.
\end{agentmessage}

%\vspace{-0.1cm}

% SQL agent tool call
\begin{agentmessage}{toolcallcolor}{SQL agent Tool Call: save\_memory}
\begin{lstlisting}[style=sqlcode]
{
    'query_string': 'identify district with second-highest crimes in 1995',
    'knowledge_string': 'SELECT district_id FROM district ORDER BY A15 DESC 
LIMIT 1 OFFSET 1;',
    'memory_type': 'similar_subtask'
}
\end{lstlisting}
\end{agentmessage}

%\vspace{-0.1cm}

% SQL agent tool response
\begin{agentmessage}{toolresponsecolor}{SQL agent Tool Response: save\_memory}
Saved.
\end{agentmessage}

%\vspace{-0.1cm}

% SQL agent tool call
\begin{agentmessage}{toolcallcolor}{SQL agent Tool Call: save\_memory}
\begin{lstlisting}[style=sqlcode]
{
    'query_string': 'join client and district tables by district_id',
    'knowledge_string': 'SELECT * FROM client INNER JOIN district ON 
client.district_id = district.district_id;',
    'memory_type': 'similar_subtask'
}
\end{lstlisting}
\end{agentmessage}

%\vspace{-0.1cm}

% SQL agent tool response
\begin{agentmessage}{toolresponsecolor}{SQL agent Tool Response: save\_memory}
Saved.
\end{agentmessage}

%\vspace{-0.1cm}

% SQL agent tool call
\begin{agentmessage}{toolcallcolor}{SQL agent Tool Call: save\_memory}
\begin{lstlisting}[style=sqlcode]
{
    'query_string': 'gender representation in client table',
    'knowledge_string': "In the client table, gender is represented by single 
characters ('M' for male, 'F' for female).",
    'memory_type': 'database_fact'
}
\end{lstlisting}
\end{agentmessage}

%\vspace{-0.1cm}

% SQL agent tool response
\begin{agentmessage}{toolresponsecolor}{SQL agent Tool Response: save\_memory}
Saved.
\end{agentmessage}

%\vspace{-0.1cm}

% SQL agent tool call
\begin{agentmessage}{toolcallcolor}{SQL agent Tool Call: save\_memory}
\begin{lstlisting}[style=sqlcode]
{
    'query_string': 'sort districts by number of crimes in 1995',
    'knowledge_string': 'To find a district based on crime rates, use the A15 
column in the district table. Order the districts by this column in descending 
order to identify districts with the highest number of crimes.',
    'memory_type': 'database_fact'
}
\end{lstlisting}
\end{agentmessage}

%\vspace{-0.1cm}

% SQL agent tool response
\begin{agentmessage}{toolresponsecolor}{SQL agent Tool Response: save\_memory}
Saved.
\end{agentmessage}

%\vspace{-0.1cm}

% SQL agent to Human
\begin{agentmessage}{sqlagentcolor}{SQL agent $\rightarrow$ Human}
I have saved the most important pieces of knowledge from our 
interaction for future use. This includes how to filter clients by gender, 
identify districts with the second-highest crimes, join client and district 
tables by district\_id, and some key database facts about gender representation 
and sorting districts by crime rates.
\end{agentmessage}

% \end{document}

\twocolumn

\end{document}